\documentclass[10pt,twocolumn,letterpaper]{article}

\usepackage[pagenumbers]{iccv} 
\usepackage[accsupp]{axessibility}  
\usepackage{dsfont}
\usepackage{etoolbox}
\usepackage{color}

\usepackage{ulem}

\newif\ifshowedits

\newcommand{\addeditor}[3]{%
  \definecolor{#1color}{rgb}{#3}
  \expandafter\newcommand\csname #1\endcsname[1]{%
  \ifshowedits
    {\color{#1color} ##1}%
  \else
    {##1}%
  \fi
  }%
  \expandafter\newcommand\csname #1rmk\endcsname[1]{%
  \ifshowedits
    {\color{#1color} {\bf [#2: ##1]}}
  \fi
  }%
  \expandafter\newcommand\csname #1rpl\endcsname[2]{%
  \ifshowedits
    {\color{#1color} ##1 \sout{##2}}
  \else
    {##1}
  \fi
  }%
}

\newcommand{\createtextvar}[1]{
  \expandafter\newcommand\csname #1\endcsname{%
  {\text{#1}}
}%
}
\newcommand{\mycomment}[1]{}

\newcommand{\calB}{{\cal B}}

\newcommand{\calL}{{\cal L}}

\newcommand{\calP}{{\cal P}}

\newcommand{\calS}{{\cal S}}

\addeditor{vincent}{VGB}{0.0, 0.5, 0.0}
\addeditor{steve}{SB}{0.5, 0.5, 0.0}
\addeditor{florian}{FC}{0.7, 0, 0.5}
\addeditor{loïc}{LB}{0, 0.5, 0.5}
\addeditor{loic}{LB}{0, 0.5, 0.5}
\addeditor{doriand}{DP}{1, 0., 0.7}
\addeditor{doreduce}{DRed}{1, 0., 0.}
\showeditsfalse

\usepackage{colortbl}
\usepackage{multirow}
\usepackage{algorithm}
\usepackage{algpseudocode}

\definecolor{iccvblue}{rgb}{0.21,0.49,0.74}
\usepackage[pagebackref,breaklinks,colorlinks,allcolors=iccvblue]{hyperref}
\usepackage{amssymb}
\usepackage{pifont}
\def\paperID{5160} 
\def\confName{ICCV}
\def\confYear{2025}

\title{DiSCO-3D : Discovering and segmenting Sub-Concepts from Open-vocabulary queries in NeRF}

\author{Doriand Petit\textsuperscript{\tiny 12}\and{}Steve Bourgeois\textsuperscript{\tiny 1}\and{}Vincent Gay-Bellile\textsuperscript{\tiny 1}\and{}Florian Chabot\textsuperscript{\tiny 1}\and{}Loïc Barthe\textsuperscript{\tiny 2}\and\textsuperscript{\tiny 1} Université Paris-Saclay, CEA, List, F-91120, Palaiseau, France, {\tt first.last@cea.fr} \and
\textsuperscript{\tiny 2} IRIT, Université de Toulouse, CNRS, France, {\tt first.last@irit.fr} 
}

\begin{document}

\twocolumn[{
\maketitle
\begin{center}
    \captionsetup{type=figure}
     \includegraphics[width=0.75\textwidth]{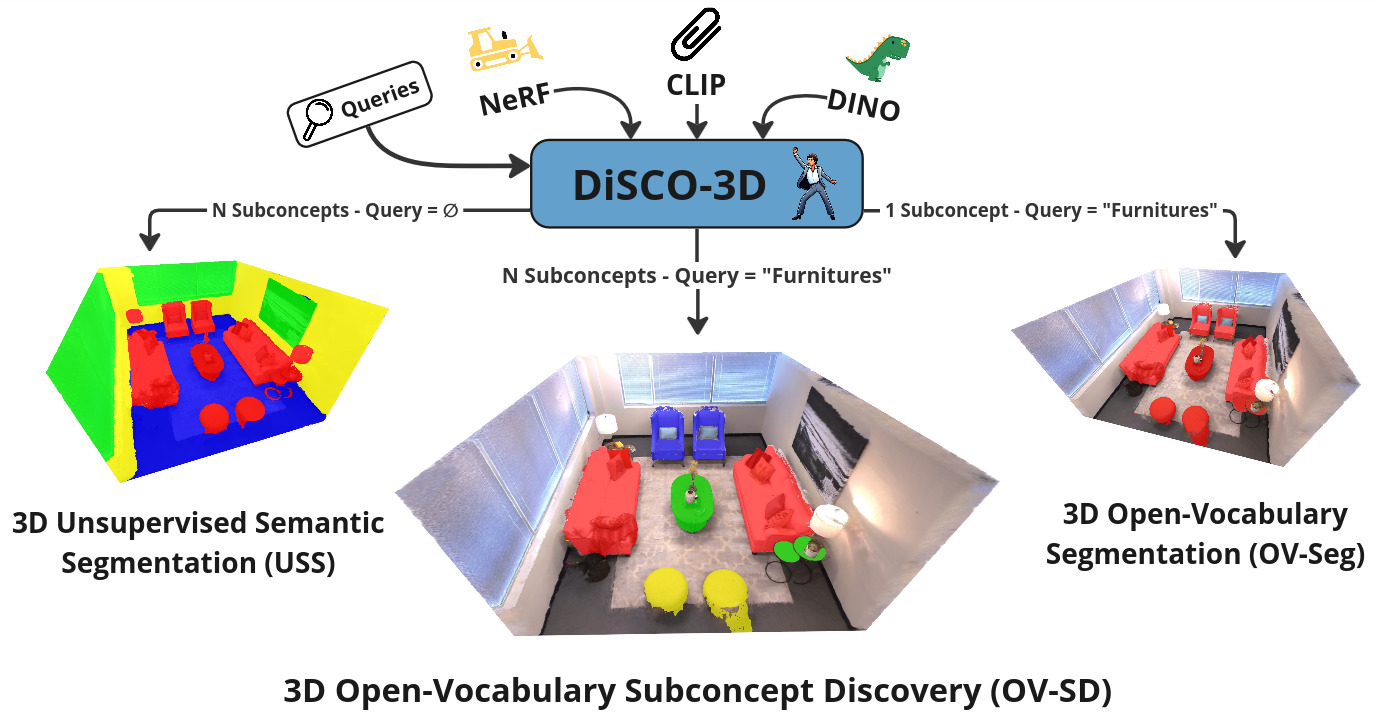} 
    \captionof{figure}{
        We introduce the 3D Open-Vocabulary Sub-concepts Discovery (OV-SD) paradigm, which aims to provide a 3D semantic segmentation adapted to both the scene (semantic classes are discovered from the scene content) and user queries (semantic classes should be semantically related to the queries, i.e. sub-concepts of queries). DiSCO-3D is the first solution to address this challenge and stands out for its versatility, as it covers 3D OV-SD’s edge cases: Open-Vocabulary Segmentation and Unsupervised Semantic Segmentation.
    }
    \label{fig:overview}
\end{center}
}]

\maketitle

\begin{abstract} 

3D semantic segmentation provides high-level scene understanding for applications in robotics, autonomous systems, \textit{etc}.
Traditional methods adapt exclusively to either task-specific goals (open-vocabulary segmentation) 
or scene content (unsupervised semantic segmentation). We propose DiSCO-3D, the first method addressing the broader problem of 3D Open-Vocabulary Sub-concepts Discovery, which aims to provide a 3D semantic segmentation that adapts to both the scene and user queries. We build DiSCO-3D on Neural Fields representations, combining unsupervised segmentation with weak open-vocabulary guidance. Our evaluations demonstrate that DiSCO-3D achieves effective performance in Open-Vocabulary Sub-concepts Discovery and exhibits state-of-the-art results in the edge cases of both open-vocabulary and unsupervised segmentation.


\end{abstract}    
\section{Introduction}
\label{sec:intro}


3D semantic segmentation~\cite{mo2022review} aims to decompose a 3D scene based on the semantic meaning of its components. This process provides a representation that emphasizes the main concepts, or semantic classes, within the scene, without distinguishing between object instances. Such high-level representations are essential in many perception applications across diverse fields, including autonomous vehicles \cite{feng2020deep}, robotics \cite{hurtado2022semantic} and medical image analysis \cite{asgari2021deep}.

In practice, however, multiple semantic decompositions are appropriate for any given scene. The suitability of a particular decomposition depends on how well it preserves relevant information for a specific downstream task. This means that the semantic segmentation should be adapted to both the content of the scene and the task's requirements.

Adaptation to the downstream task should involve not only providing the relevant semantic classes for the task, but also excluding irrelevant ones since, as underlined by ~\citet{eftekhar2023selective}, they would behave as distractors. Scene adaptation, on the other hand, should require that the output semantic classes provide the most fine-grained semantic description of the scene, while also excluding classes absent in the scene.
To illustrate the difference, if a scene contains a television, a hammer and a screwdriver, and the task requires the use of tools, task adaptation implies to ignore the "television" class while adaptation to the scene implies to provide the two classes "hammer" and "screwdriver" instead of a single class "tools" or a list of all existing tools whether or not they actually are in the scene.  
However, supervised approaches~\cite{zhou2022supervised} as well as more recent Open-Vocabulary methods (OV-Seg)~\cite{peng2023openscene,kerr2023lerf,liu2023weakly,shi2024language} focus on adapting segmentation to a specific downstream task by requiring users to specify task-relevant classes, whereas recent works on 3D Unsupervised Semantic Segmentation (USS)~\cite{zhang2023growsp,liu2024u3ds3} focus exclusively on adapting the semantic classes to the scene through label-free decomposition. To our knowledge, no solution provides both adaptations.




We introduce \textit{3D Open-Vocabulary Sub-concepts Discovery} (3D OV-SD), which involves providing the most relevant segmentation of a 3D scene regarding its content and a downstream task defined through a user query. We propose DiSCO-3D as the first solution, consisting in plugging into a Neural Field \cite{mildenhall2021nerf} representation an 
USS module partially supervised by 
an OV-Seg.
As illustrated in \autoref{fig:overview}, DiSCO-3D not only addresses OV-SD but also generalizes to its edge cases: 3D 
OV-Seg (when queries target a single sub-concept) and 3D 
USS (when no query is provided).
Our main contributions are:
\begin{enumerate} 

    \item We introduce 
    3D OV-SD, a new 3D semantic segmentation task 
    providing adaptive segmentations based on scene context and user-defined queries. 
    We also propose a 
    quantitative benchmark by extending Replica's semantic classes and providing a suitable evaluation protocol.
    \item We present DiSCO-3D, the first method designed to solve the 3D OV-SD problem, combining Unsupervised Semantic Segmentation with Open-Vocabulary Segmentation guidance to serve as a direct plug-in to NeRF.  
 
    \item We evaluate DiSCO-3D on both real and synthetic data, demonstrating better performance than hand-designed naive baselines on the proposed OV-SD task and experimentally show that our solution produces state-of-the-art performances on the OV-SD edge cases of NeRF Open-Vocabulary Segmentation and Unsupervised Semantic Segmentation, highlighting its versatility.

\end{enumerate}


\section{Related Works}
\label{sec:sota}

\textbf{Unsupervised Semantic Segmentation.}
Due to the difficulty of obtaining large annotated datasets, the unsupervised paradigm has attracted attention for image semantic segmentation. 
Recently, 2D USS approaches have adopted self-supervised \steverpl{}{, semantically-aware} pre-trained models, such as DINO \cite{caron2021emerging}, as input for deep clustering modules. In particular, STEGO \cite{hamilton2022unsupervised} inspired a range of techniques by revealing the correlation between unsupervised network features and true semantic labels. This line of research has recently expanded with methods like ACSeg \cite{li2023acseg}, EAGLE \cite{kim2024eagle}, and SmooSeg \cite{lan2024smooseg}, which focus on online clustering of pixel-level features, typically by contrasting these features into easily classifiable groups. Apart from 2D USS ideas, some recent methods \cite{zhang2023growsp,liu2024u3ds3} focus on performing unsupervised semantic segmentation directly on 3D point clouds. These methods demonstrate an increasing interest in transferring label-free semantic segmentation to 3D. 
While USS methods adjust segmentation to the scene's content, they are independent of downstream tasks, making the output classes poorly suited for follow-up applications. \\ 
\newline
\textbf{3D Open-Vocabulary Segmentation.} 2D Open-Vocabulary segmentation methods rely on the use of pre-trained vision-language models such as CLIP~\cite{radford2021learning}, sharing a feature space for both image and text encodings. 
Although the original models produce per-image embeddings, several solutions focus on computing pixel-wise features for precise segmentation~\cite{li2022language,ghiasi2022scaling,ding2023maskclip,wysoczanska2024clipdino}. Due to the high cost of 3D data acquisition and annotation, developing 3D foundation models is challenging, which has led to extensive research on applying 2D Vision-Language models for 3D open-vocabulary segmentation. Many approaches have indeed proposed to distill various 2D foundation models~\cite{radford2021learning,caron2021emerging,oquab2023dinov2,kirillov2023segment,zou2023generalized,ghiasi2022scaling} into various 3D representations ranging from point clouds \cite{peng2023openscene} to more recent Neural Radiance Fields \cite{mildenhall2021nerf} and Gaussian Splatting \cite{kerbl3Dgaussians}. 
More specifically, NeRF-based distillation into so-called \textit{feature fields} are particularly well-known as the ray-based nature of NeRF is highly compatible with feature distillation. 
Notably, many works distill various image encoders into their models for different semantic applications~\cite{kobayashi2022decomposing,chen2023interactive,tschernezki2022neural,shen2023distilled,liu2024sanerf,kim2024garfield}, ranging from semantic segmentation~\cite{kobayashi2022decomposing,tschernezki2022neural} to open-vocabulary segmentation~\cite{kerr2023lerf,liu2023weakly}. 
Some methods, such as LeRF~\cite{kerr2023lerf} or LEGaussians~\cite{shi2024language}, combine different feature fields into a single model to leverage the strengths of each encoder. 
\steverpl{
    Although 3D open-vocabulary segmentation methods decompose scenes based on user queries, their semantic labels are limited to these concepts. 
    Our method rather automatically discovers 
    sub-concepts, offering a richer scene description and flexibility for real-world applications.
}{
    Although 3D open-vocabulary segmentation methods provide a 3D scene decomposition based on user-defined queries, each semantic label is limited to the queried concepts, which heavily restricts the flexibility for real-world applications. On the other side, our method takes into account the content of the scene to automatically discover the sub-concepts related to easier-to-nominate queries, and provides consequently a richer description of the scene.
}

\section{DiSCO-3D}

\begin{figure*}[t]
    \includegraphics[width=\textwidth]{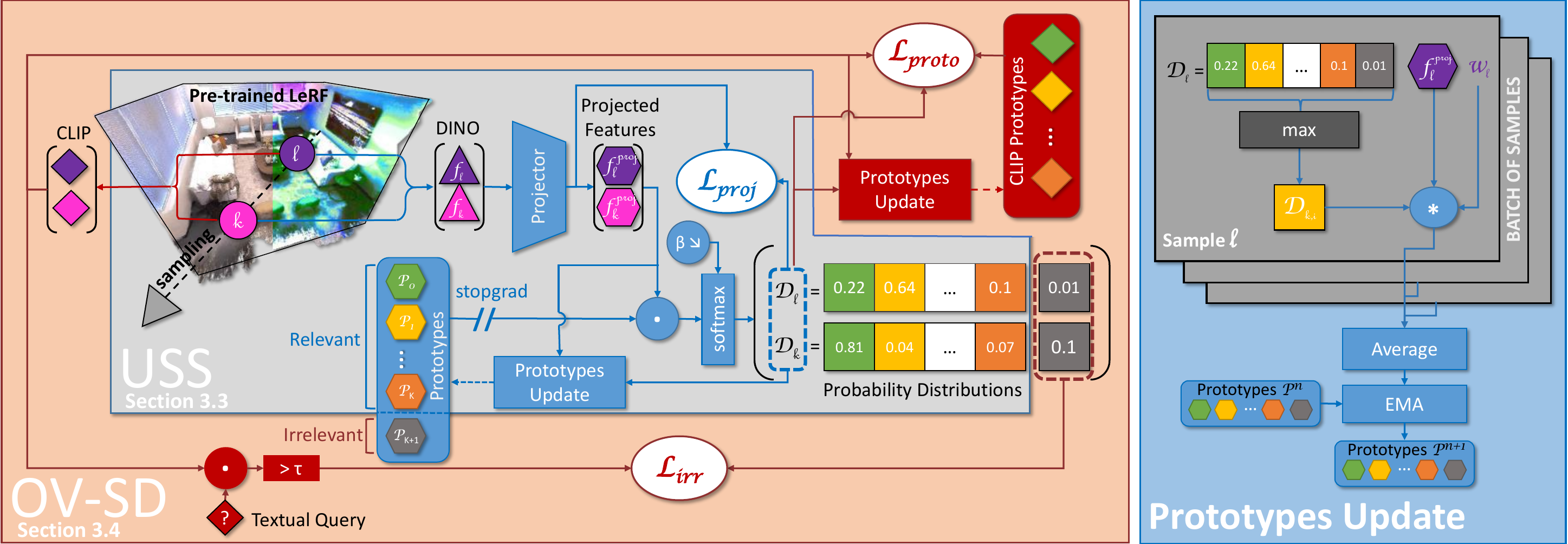}
    \caption{\textbf{Overview of DiSCO-3D for a LeRF Feature Field.} DiSCO-3D inputs pairs of features from 3D samples into a projector network learnt to accentuate semantic disparities. Those projected features are then classified by comparing them to class-specific prototypes (\autoref{sec:USS}). Thoses prototypes are updated each epoch using an EMA with the projected features. A user query can be used to supervise the projector by encouraging the prototypes to be divided into either relevant or irrelevant classes, enabling the semantic segmentation of only task-relevant sub-concepts (\autoref{sec:subconceptDiscovery}).}
    \label{fig:overview_method}
\end{figure*}

\subsection{Problem Statement and Overview}
\label{sec:pb}

The objective of our method is to provide a solution to the previously introduced 3D Open-Vocabulary Sub-concepts Discovery problem, specialized to the case of Neural Field \cite{mildenhall2021nerf} representations. 
As illustrated in \autoref{fig:overview}, it consists in providing a 3D semantic segmentation of the scene related to one or several user queries without explicit nomination of each of the semantic classes. This task requires both to understand what's relevant in a scene relative to user queries and being able to semantically cluster these relevant objects without any additional information on the user requirements. Natural solutions would include the successive works of both open-vocabulary understanding of the scene to decipher query relevancy across the scene and unsupervised semantic segmentation to propose scene-adapted semantic decomposition (in any order).
However, performing these two sub-tasks successively gives sub-optimal results, as will be demonstrated in~\autoref{sec:sub} and this is why we build DiSCO-3D to perform them simultaneously: an Unsupervised Semantic Segmentation module based on prototypes that automatically discovers semantic classes in the scene, alongside a parallel mechanism leveraging open-vocabulary segmentation to guide the USS toward sub-concepts related to one or more user queries. Those modules are connected together through the use of a shared architecture, as illustrated in \autoref{fig:overview_method}. 

To reach this objective, our solution relies on a pre-trained frozen Neural Field representation of the 3D scene containing both a queryable representation (eg. an Open-Vocabulary field \cite{radford2021learning,ghiasi2022scaling}) and a spatially precise semantic representation \cite{caron2021emerging,oquab2023dinov2} (called feature fields). For ease of understanding, we first consider the specific case of a pre-trained LeRF \cite{kerr2023lerf} which includes jointly a multi-scale CLIP \cite{radford2021learning} pyramid and a DINO \cite{caron2021emerging} feature field. However, other feature fields can be used, as discussed in \autoref{sec:extensions}.

In the following, we present our method in three parts. After explaining some preliminaries information in \autoref{sec:preli}, we first introduce an extension to 3D Unsupervised Semantic segmentation method adapted for Neural Fields (\autoref{sec:USS}). We then extend this approach to perform open-vocabulary sub-concepts discovery by incorporating open-vocabulary guidance into the USS process (\autoref{sec:subconceptDiscovery}). 
Finally, we introduce several extensions to handle more complex queries and diverse semantic representations (\autoref{sec:extensions}). 

\subsection{Preliminaries}
\label{sec:preli}

\textbf{NeRF and Feature Fields.} 
Neural Radiance Fields \cite{mildenhall2021nerf} (NeRFs) are learnable neural networks (possibly coupled with multi-resolution feature hashgrids \cite{mueller2022instant}) overfitted to individual scenes, which output density ($\sigma$) and color ($c$) from any 3D position and view direction queries. A 2D pixel color $\hat{C}$ is recovered by sampling points along a ray cast from the corresponding posed image and compositing them via volume rendering: $\hat{C}(r) = \sum_{i=0}^{N-1} w_i c_i$, with $w_i = T_i(1-exp(-\sigma_i \delta_i))$ (which we denote as the density weights) and $T_i = exp(\sum_{j=0}^{i-1}\sigma_j \delta_j)$, $c_i$ is the color of sample $i$ and $\delta$ is the distance between consecutive samples. The scene is optimized by minimizing the MSE loss $\mathcal{L}_{rgb} = ||\hat{C}(r) - C(r)||^2$ between rendered and ground truth colors.
Feature fields are trained similarly by replacing RGB color with d-dimensional features, optimizing the model via comparison between NeRF rendered features and feature maps from pre-trained image encoder. For example, LeRF~\cite{kerr2023lerf} jointly learns a multi-scale CLIP pyramid (using image patches) and DINO feature fields inside a single model (with joint feature grids but separate decoders). 

\subsection{Unsupervised Semantic Segmentation for LeRF}
\label{sec:USS}

To the best of our knowledge, USS approaches have never been adapted to the continuous NeRF representation. Hence, to perform 3D unsupervised semantic segmentation from a LeRF representation, we draw inspiration from well-known prototypes-based 2D methods \cite{hamilton2022unsupervised,li2023acseg,kim2024eagle,lan2024smooseg} which focus on clustering semantic features from pre-trained vision models like DINO. Similarly to these approaches, our solution relies on  
learning a non-linear projector (see supplementary material for architecture details) that maps the scene DINO features obtained from LeRF onto a new latent space where the projected features are agglomerated around cluster centroids (also named prototypes) which each describe a semantic class.

Given a sample $k$, the probability distribution of class assignment $D_k$ is computed as the softmax (augmented with an additional $\beta$ sharpness hyperparameter) of the cosine similarity between the projection $f^{proj}_k$ of the DINO feature $f_k$ and each of the $N$ prototypes $\calP_i$:
\begin{equation}
    D_k = \text{softmax}(f^{proj}_k\cdot\mathcal{P}_i/\beta, i\in[1,N])
    \label{eq:Ds}
\end{equation} 

Regarding the training process, this one is achieved per batch of $\calB$ DINO features which we obtain by casting rays from randomly sampled pixels across all available posed images, and then sampling the scene along these rays, following usual NeRF pipelines. While the projector is optimized through standard back-propagation, the prototypes are updated using a different strategy. At a given epoch $n$, each prototype $\calP_i$ is updated using the weighted average of the post-projection features $f^{proj}_k$ whose associated samples $\calS_i$ are classified as the class $i$, the prediction confidence $D_{k,i}$ (i.e. the $i^{th}$ component of $D_k$) serving as weight. Moreover, unlike image or point cloud representations, the relevance of the DINO feature field varies depending on the 3D position within the scene. Specifically, if a sample’s 3D location is in free space or inside an object, the corresponding DINO feature value becomes meaningless and should not significantly contribute to the training process. Because the relevance of 3D samples along a given ray is determined by their density weights $w_i$ (see \autoref{sec:preli}), we also adjust sample contribution to the prototypes update according to these weights. We thus transform the weighted average into a two-fold weighted average of $w_{k}D_{k}$ as illustrated in \autoref{fig:overview_method}. To ensure stability during training, we apply an Exponential Moving Average (EMA) across epochs, resulting in the following update process for all $i \in [1,N]$ :
    
\begin{equation}
\mathcal{P}^{n+1}_{i} = \alpha \mathcal{P}^n_{i} + (1- \alpha) \frac{ \sum_{k\in S_i} w_k D_{k,i} f_k}{\sum_{k\in S_i } w_k D_{k,i}},  \forall i \in [1,N] 
\label{eq:EMA}
\end{equation}

The projector’s supervision presents two main challenges. The first is ensuring the projection maintains the semantic consistency of the DINO features, preserving the distance relationships between them (i.e., features close in DINO space should remain close in the output space, and vice versa). The second challenge is achieving well-separated clusters with sharp probability distributions.

To address the first challenge, we incorporate a loss, denoted $\calL_{proj}$, designed to maintain the relationships between DINO features and their projected counterparts. This loss, commonly known as correlation loss~\cite{hamilton2022unsupervised}, smoothness loss~\cite{lan2024smooseg} or correspondence distillation loss~\cite{kim2024eagle}, uses pairs of samples to supervise the projector (using a \textit{stopgrad} operation on the prototypes) by encouraging pairs that are close in DINO space (i.e. closer than a fixed hyperparameter $b$) to have similar probability distributions  while encouraging pairs that are distant to exhibit more divergence in their distributions: 
\begin{equation}
\calL_{proj} = \frac{1}{\mathcal{B}} \sum_{k,l}^{\mathcal{B}} (\frac{f_k \cdot f_l}{||f_k|| ||f_l||} - b) (1 - D_k \cdot D_l)
\label{eq:lproj}
\end{equation}

Although this loss maintains semantic consistency, it does not prevent the probability distribution of a projected feature of the scene from being uniform or relatively smooth over the semantic classes. 
To solve this issue, unlike other USS methods relying on additional losses \cite{lan2024smooseg}, we propose to enforce a progressive sharp agglomeration of the samples around their associated prototypes by introducing a scheduled linear decaying of the $\beta$ parameter (defined in \autoref{eq:Ds}) to progressively separate the clusters.\\

\subsection{Open-vocabulary Guidance for Sub-concepts Discovery } \label{sec:subconceptDiscovery}

We build on the previously introduced LeRF USS segmentation approach to address the core challenge of our paper: 3D Open-Vocabulary Sub-concepts Discovery. 
Let's first consider the scenario of a unique query.\\
\textbf{Discovering Query-Relevant Sub-concepts.}
We aim in the following to discover and segment $N_q$ sub-concepts of a scene related to a user query represented as a CLIP embedding $q$.
Since these sub-concepts are not specified by the user and depend on the scene, it is not possible to provide a supervision for each of them. On the other hand, the irrelevant semantic parts of the scene are implicitly defined by the query. We thus propose, as illustrated in \autoref{fig:overview_method}, to extend the previously presented 3D USS method by adding an additional \textit{irrelevant} semantic class with $N_{irr}$ associated prototypes (corresponding to index $N_q+1$ to $N_q+N_{irr}$). The projector is then supervised by an additional loss function $\calL_{irr}^q$: 
\begin{equation}
    \calL_{irr}^q = \frac{1}{\# \overline{M_q}} \sum_{k\notin M_q}     (D_k \cdot H_q^T) + \frac{1}{ \# M_q} \sum_{k\in M_q} (1-D_k \cdot H_q^T)   
    \label{eq:LossIrrelevant} 
\end{equation} 
where $M_q$ is the binary mask representing the open-vocabulary segmentation of the CLIP field based on the embedding $q$, obtained by applying a threshold $\tau$ to the batch cosine similarity between $q$ and $f^{CLIP}$; $\overline{M_q}$ is its complement; and $H_q = [\underbrace{1,1,...,1}_{N_q \text{first terms}},\underbrace{0,0,...,0}_{N_{irr} \text{last terms}}]$ is the one-hot vector associated with the relevant semantic class.
This loss uses the CLIP field to supervise the DINO projector by maximizing the probabilities of irrelevant classes for samples unrelated to the query, and minimizing them for other samples. Combined with $\calL_{proj}$ applied on relevant classes, it enables the model to focus on discovering and segmenting relevant sub-concepts.\\
\textbf{Semantic Recovery and CLIP-Guided Regularization.}
Because the projector is specific to each training process, the prototypes are defined on an arbitrary feature space rather than a scene-agnostic foundation model's embedding space. This limits the ability to fully leverage the results, allowing only the use of the sub-concepts segmentation maps but without the ability to understand them.

To handle this problem, we propose to associate a $\calP_{i}^{CLIP}$ embedding to each prototype $\calP_i$. 
These are initialized as zero embeddings and are updated using the same EMA process as the base prototypes, with the weighted mean of the CLIP embeddings of the batch samples associated with the corresponding class, as illustrated in \autoref{fig:overview_method}.

 These characteristic embeddings can be used for various applications, notably \textit{a posteriori} concept retrieval as illustrated in \autoref{sec:sub}. Moreover, by design, this generic prototype definition can be applied to any available feature field (e.g., DINO in LeRF) using the same approach, enabling further semantic understanding.
 
These CLIP prototypes can also be leveraged during optimization to enhance semantic segmentation performance. Since the projector uses DINO features as input which tends to produce over-segmented features (i.e. describing object parts rather than entire objects), we regularize the model using CLIP semantics (which are more object-consistent) by introducing a final loss term based on the CLIP prototypes. This loss, denoted as $\calL_{proto}$, drives the projected DINO features closer to the prototype whose CLIP embedding is most similar to the sample's CLIP embedding:
 
 \begin{equation}
     \calL_{proto} = \frac{1}{\calB}\sum_{k}^{\calB} (1-D_k \cdot H_k^T)
      \label{eq:LossReg} 
 \end{equation}
   
 where $H_k$ is a one-hot tensor defined such that the one is at $ \underset{i\in[1,N]}{\mathrm{argmax}}\, (f^{CLIP}_k.\calP^{CLIP}_i)$ for every sample of the batch. 
 \\
 


\subsection{Method extensions } \label{sec:extensions}

\begin{figure*}[t]
    \includegraphics[width=\textwidth]{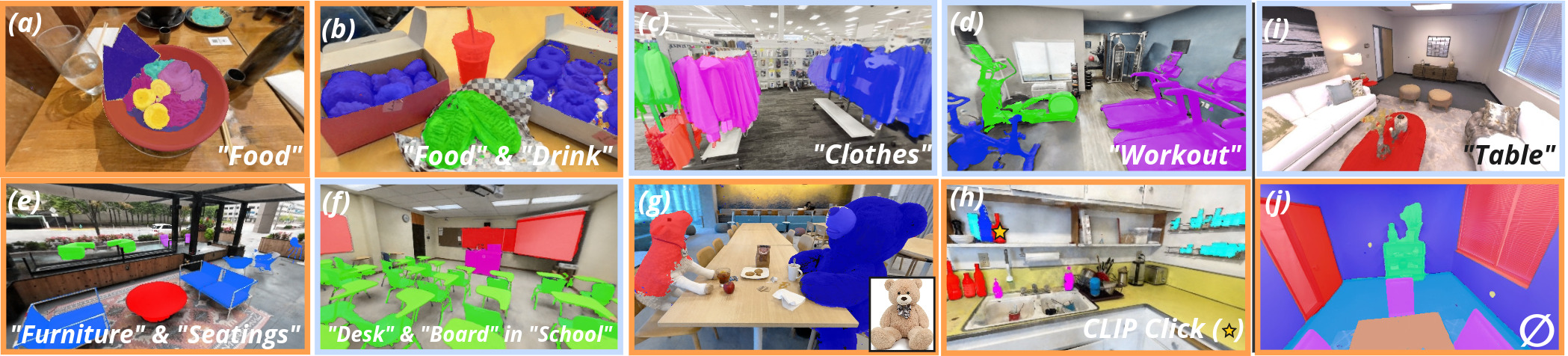}    
    \caption{\textbf{DiSCO-3D Qualitative Evaluation for OV-SD.} We present results for various queries, scenes (which originate from \cite{kerr2023lerf,ling2024dl3dv,replica19arxiv}) and feature fields (LeRF in orange and OpenNeRF blue). (b), (e) and (f) illustrate multiple queries, resp. disjoint, overlapping and nested, (g) a visual query encoded with CLIP and (h) a CLIP feature obtained by a user click as query. Finally, (i) and (j) are OV-SD edge cases, where (i) has 1 sub-concept (OV-Seg) and (j) has no user query (USS).}
    \label{fig:resu}
\end{figure*}

\textbf{Multiple and complex queries.}
So far, we have described the method with a single query for clarity. However, DiSCO-3D can process multiple simultaneous queries $Q=\{q_i,i\in[1,K]\}$, as long as we define \textit{a priori} which prototypes are relevant for each query. While the losses $\calL_{proj}$ and $\calL_{proto}$ remain unchanged, a loss $\calL_{irr}^{q_i}$ is added for each query $q_i$ following \autoref{eq:LossIrrelevant}. Each of these losses is guided by a unique one-hot vector $H_{q_i}$ that defines the relevant prototypes for each query. Notably, it is filled with ones for the defined relevant prototypes and filled with zeros elsewhere (both irrelevant and non-overlapping prototypes from other queries).
This formulation allows full flexibility, supporting overlapping, disjoint, or nested queries without additional constraints.\\
\textbf{Extending to other Features Fields.} Although we present our method using a pre-trained LeRF as input, DiSCO-3D is compatible with a wide range of feature fields (and their combinations) as long as two conditions are met. First, the projector requires at least one spatially precise feature field to perform segmentation (e.g., dense encoders). Second, the scene must be represented by at least one feature type that can be compared to a query. Given these conditions, the input 3D representations and query modalities can vary widely—from a single feature field satisfying both requirements (e.g. OpenSeg in \autoref{sec:sub}) to alternative inputs such as user clicks, as demonstrated in \autoref{fig:resu}.


\section{Experimental evaluations}
\label{sec:exps}
After introducing some implementation and evaluation details in~\autoref{sec:impl}, we first present evaluations on the novel Open-Vocabulary Sub-concepts Discovery problem with a dedicated benchmark in~\autoref{sec:sub}. Then, we successively propose experiments for the edge cases of Open-Vocabulary Segmentation and Unsupervised Semantic Segmentation in~\autoref{sec:specific}. Additional details on hyperparameters, evaluation protocols and baselines can be found in the supplementary materials, as well as ablative experiments and analysis on DiSCO's limitations.

\subsection{Implementation and evaluation details}
\label{sec:impl}

We implemented our method in the Nerfstudio \cite{tancik2023nerfstudio} framework and every evaluation is based on the same Nerfacto model, a grid-based NeRF method coupled with several Mip-NeRF-360 \cite{barron2022mip} improvements. Regarding the feature fields, we follow LeRF's implementation and add a set of grid shared amongst features with a dedicated MLP decoder for each field. For quantitative evaluation, we use both LeRF's CLIP and DINO, and OpenNeRF's OpenSeg dense CLIP feature fields. Details on the architecture hyperparameters and image encoders can be found in supplementary materials.
All quantitative experiments, including DiSCO-3D and the comparative baselines, use the same pre-trained Nerfacto models and feature fields as input. All our experiments were run on the same single RTX 4090 GPU. They run for 100 epochs each, at approximately 20ms per epoch (resulting in $\sim$2s optimization per query, which can be considered fast enough for most practical applications; see sup. mat. for further discussions on DiSCO's speed).\\
    \textbf{Dataset.}
    We introduce an extension of the Replica~\cite{replica19arxiv} dataset for Open-Vocabulary Sub-concepts Discovery. This extension consists of enriching the annotations of its 8 indoor synthetic scenes with 40 semantic concepts. Each concept represents a specific grouping of Replica classes (called sub-concepts) and is designed with robotic perception in mind. To ensure diversity, we use a large language model (LLM) to generate queries spanning object categories (e.g., "furniture"), properties (e.g., "soft"), and actions (e.g., "eat"). The complete list of concepts and their sub-concepts is provided in the supplementary material. \\  
    \textbf{Protocol.}
    The following evaluation protocol is designed to be compatible with any OV-SD method and follows the standard approach of being performed on a segmented point cloud, which can typically be derived from most types of 3D representations. It assesses, for each scene and concept, the segmentation quality of discovered sub-concepts against the query-related ground-truth sub-concepts. The evaluated method takes as input the scene $S$, the textual query $q$ corresponding to a concept, and a collection of 3D points $P$. It should output a set of embedding (e.g. CLIP) representing the discovered sub-concepts of $q$ and should classify each point of $P$ with at most one of these embeddings (in DiSCO, these relate to $\calP_{CLIP}$ associated to each prototype). 
    Segmentation quality is evaluated by first matching discovered sub-concepts to the dataset-defined sub-concepts (we match predictions with all of the scene's classes) using embeddings distances. This enables comparison with the ground-truth query segmentation to compute classic segmentation metrics: Mean Accuracy (mAcc) and mean Intersection over Union (mIoU). Since these metrics do not penalize the presence of unrelated predicted sub-concepts (i.e. false positive classes not matched to any ground-truth sub-concepts), we also use the standard Panoptic Quality~\cite{kirillov2019panoptic} (PQ) metric, replacing the notion of instances with sub-concepts. \\
    \textbf{3D Point Cloud Conversion.} For NeRF-based methods such as DiSCO, in order to obtain 3D point cloud predictions, we follow OpenNeRF’s protocol and render the semantic class distribution image (in DiSCO-3D, we compute $D_i$ for each pixel $i$) for each supervision viewpoint and back-project it onto the 3D point cloud. Probabilities for each 3D point are then aggregated across viewpoints, with the final class assigned via an \textit{argmax} operation.

\subsection{Open-Vocabulary Sub-concepts Discovery} 
\label{sec:sub}

    \subsubsection{Evaluated methods.}  Since no OV-SD baselines exist yet, we design and evaluate two naive baselines alongside DiSCO. Both of these baselines share the DiSCO-3D architecture and the input feature fields. The first one begins by performing open-vocabulary segmentation to identify relevant regions (i.e. regions where the CLIP similarity with the query is above a fixed threshold) and then does USS on those regions. The second baseline runs USS on the full scene and then filters out irrelevant classes by thresholding the CLIP similarity of each USS class to the query (using their CLIP prototypes). Notice that the only difference between DiSCO-3D and those baselines relies on the fact that DiSCO-3D achieves USS and OVSeg jointly whereas the latters achieve it successively. We also create two additional naive baselines by replacing the USS part by K-Means.
    All the methods use the same hyperparameters and especially, we fix the number of prototypes $N=10$ for all queries (as no concept query exceeds $9$ ground-truth sub-concepts). 
        
    \subsubsection{Results}
     Quantitative results are reported in~\autoref{tab:ovsd} (in $\calP_{CLIP}$ columns). We notice that using DiSCO-3D always \steverpl{overperforms}{performs better than any of} the naive baselines, which demonstrates the interest of performing jointly USS and OVSeg. More specifically, considering the PQ, mIoU and mAcc averaged on both LeRF and OpenNeRF, we obtain respectively an increase of $+72\%$, $+71\%$ and $+42\%$  against USS→OVS, and $+47\%$, $+22\%$ and $+44\%$ against OVS→USS. However, \steverpl{the benchmark is still far from being saturated, showing the difficulty of the task and the room for future improvements.} {it should be noted that metrics remain quite low, underlying the fact that the task and benchmark are challenging, allowing further improvements.} 
     
     We also display some qualitative examples in~\autoref{fig:resu} across various scenes (both indoor and outdoor from various datasets \cite{kerr2023lerf,ling2024dl3dv,replica19arxiv}), feature fields (LeRF and OpenNeRF),  types of queries (textual, visual and user clicks) and queries complexity (multiple queries at once). \steverpl{Although}{We notice that, although} the overall results are strong, some minor inaccuracies (e.g., armrests in (e)) or missed ambiguous detections (e.g., workout device at the back in (d)) can still be observed.

    

    \subsubsection{Ablations studies}
    \textbf{Accuracy of CLIP Prototypes.} To evaluate the ability of the produced CLIP prototypes to achieve semantic matching, we evaluate the OV-SD performance by replacing the prototypes matching by a \steverpl{ground-truth aware matching}{spatial matching which considers the best possible correspondence between the discovered and ground-truth sub-concepts }(using the Hungarian algorithm between ground-truth and predicted 3D segmentation masks). Results are reported in~\autoref{tab:ovsd} ($Hungarian$ columns).
    
    
    
\begin{figure}[b]
    \includegraphics[width=\columnwidth]{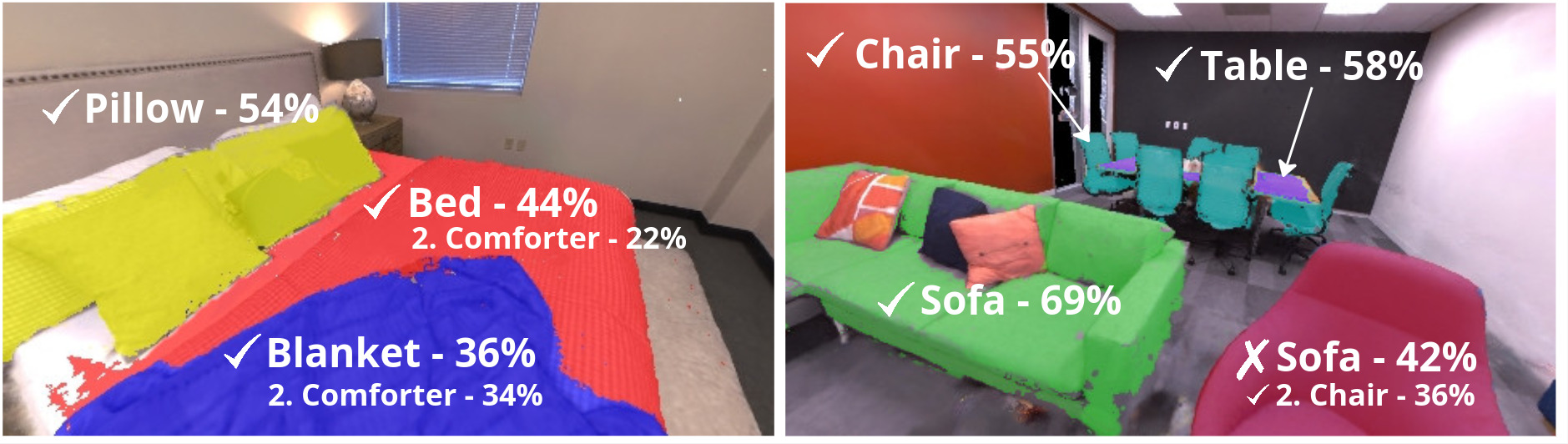}    
    \caption{\textbf{Linking Sub-concepts to \textit{a posteriori} Textual Classes.} The queries of the left and right images are respectively "Sleep" and "Furniture". By comparing each CLIP prototype to Replica's semantic classes encoded with CLIP, DiSCO-3D is able to choose the most relevant class to describe each sub-concept.}
    \label{fig:resuLabel}
\end{figure}

    \doriand{ 
        Considering the PQ, mIoU and mAcc averaged on both LeRF and OpenNeRF, we observe respectively an increase of $23\%$, $18\%$ and $42\%$ when the optimal GT-\steverpl{aware}{based} matching is used. In practice, we compute that approximately $76\%$ of the matching remains unchanged, while $24\%$ is reassigned to a new ground-truth sub-concept. As illustrated in \autoref{fig:resuLabel}, this $24\%$ is usually related to sub-concepts with close semantic (e.g. "blanket" and "comforter") or ambiguous annotations (e.g. the armchair is \steverpl{annnotated as "chair" whereas DiSCO confidence better reflects the ambiguity with the  "sofa" and "chair" sub-concepts}{ predicted as a "sofa" rather than a "chair"}). It underlines that many discovered sub-concepts have descriptive CLIP prototypes which may be sometimes ambiguous due to the nature of OV-SD.
    }
    
     
    Finally, we observe that the difference of performances between DiSCO and the baselines is not related to the use of CLIP prototypes. Indeed, averaged on both feature fields on the GT matching, DiSCO provides an increase of $+53\%$, $+30\%$ and $+23\%$ against USS→OVS for PQ, mIoU and mAcc, and $+45\%$, $+19\%$ and $+49\%$ against OVS→USS. We also notice that replacing our USS with K-Means in the naive baselines outputs mostly worse performances, highlighting the interest of our architecture choices. \\
    \textbf{Sensitivity to Number of Prototypes and influence of $\calL_{proto}$.}
    Following 2D USS evaluations~\cite{li2023acseg}, we study the impact of the predefined number of prototypes on DiSCO's performance in~\autoref{tab:proto}. We vary $N$ from $N_{GT}$ (the number of ground-truth sub-concepts in a specific scene for a specific query) to $N_{GT} + 20$ and report both segmentation performance and the actual number of prototypes used in practice, with and without $\calL_{proto}$. To facilitate interpretation, we use ground-truth aware matching.
    
    First, we observe that the complete model’s performance remains stable in both segmentation accuracy and the number of prototypes used, as long as a sufficient number of prototypes is available. Small performance drops for $N=N_{GT}$ and $N=N_{GT}+2$  likely stem from the risk of missing GT classes, as the limited number of available prototypes leaves no flexibility for some to remain unused. Adding $\calL_{proto}$ effectively regularizes the number of prototypes used, leading to improved PQ and confirming its role in optimizing prototype selection. Finally, the last column, corresponding to our main experiment with a fixed $N = 10$, shows that performance is maintained without requiring prior knowledge of the number of GT sub-concepts.

\begin{table}[tb]
\centering
\scalebox{0.64}{
    \begin{tabular}{@{\extracolsep{8pt}}ll lll lll@{}}
    \hline
        
       \multirow{2}{*}{FF} & \multirow{2}{*}{Method}  &  \multicolumn{3}{c}{$\calP_{CLIP}$}  &
          \multicolumn{3}{c}{\textit{Hungarian}}
           \\ \cline{3-5} \cline{6-8}
      &  & PQ $\uparrow$  & mIoU $\uparrow$ &mAcc $\uparrow$ & PQ $\uparrow$ & mIoU $\uparrow$ & mAcc $\uparrow$ 
        \\ 
        \hline
    
    \multirow{3}{*}{\rotatebox[origin=c]{90}{LeRF}} &   
    K-Means→OVS & - & - & - & 6.32 & 8.82 & 20.97 \\
    &  USS→OVS & 4.76 &    6.52  &  22.54 &  6.94   &  10.92  &  35.57  \\
    &  OVS→K-Means & - & - & - & 6.59 & 10.88 & 24.35 \\
    &   OVS→USS       & 5.99  & 8.71  & 21.44 & 7.48 &  10.90  & 27.11 \\
    &    DiSCO-3D  & \textbf{8.13} & \textbf{10.79} & \textbf{33.39} & \textbf{10.19}  & \textbf{12.77}  & \textbf{44.29}    \\ \hline 
     \multirow{3}{*}{\rotatebox[origin=c]{90}{\footnotesize OpenNeRF}} & 
     K-Means→OVS & - & - & - & 5.80 & 8.28 & 20.43 \\
    &  USS→OVS  &  4.97  & 6.08 & 13.98 &   6.53  & 8.67 & 23.85 \\ 
    &  OVS→K-Means & - & - & - & 6.67 & 10.46 & 23.88 \\
     & OVS→USS  &  5.47  &  8.94 &  13.56 &  6.73   & 10.58 & 22.00 \\ 
    &  DiSCO-3D    & \textbf{8.65} &  \textbf{10.82} & \textbf{19.24} &  \textbf{10.49}  & \textbf{12.69} & \textbf{29.06} \\ \hline
    \end{tabular}
}

\caption{\textbf{Quantitative Evaluation for OV-SD.} Additional metrics can be found in sup. mat. "FF" stands for feature field.}
\label{tab:ovsd}
\end{table}


\begin{table}[tb]
\centering
\scalebox{0.75}{
    \begin{tabular}{c | c | c | c | c | c | c || c}
$\calL_{proto}$& $N_{add}$  & 0 & 2 & 5 & 10 & 20 & $N=10$\\\hline

 \multirow{2}{*}{\ding{55}} & Used $N_{add}$ & -0.07 & 1.33 & 1.98 & 2.62 & 3.02 & 2.60 \\ 
  & PQ $\uparrow$  &  8.56 & 9.49  & 9.72  & 9.71  & 9.55  &  \textbf{9.77} \\\hline
  
\multirow{2}{*}{\ding{51}} & Used $N_{add}$ & -0.12 & 1.08 & 1.52 & 1.91 & 1.96 & 1.80 \\ 
& PQ $\uparrow$  &  8.53 & 9.52  & 10.06  & 10.15  & 10.12  &  \textbf{10.19} \\

    \end{tabular}
}
\caption{\textbf{Ablative on $\#$ of Prototypes} ($N = N_{GT} + N_{add}$). These are done in the \textit{Hungarian Matching} paradigm and with LeRF. The last column refers to the main experiment where the number of prototypes is fixed and does not depend on $N_{GT}$. \steve{The line "Used $N_{add}$" represent the average difference between the number of ground-truth sub-concepts and the number of  sub-concept prototypes actually used by DiSCO-3D.}}
\label{tab:proto}
\end{table}

\subsection{Edge Cases}
\label{sec:specific}

Beyond the general OV-SD task, DiSCO-3D demonstrates remarkable versatility by effectively handling its specific edge cases. It seamlessly adapts to OV-Seg, a simplified form of OV-SD where each query asks for a single sub-concept, and to USS, which can be seen as OV-SD without any user query. In the following, we evaluate DiSCO on both of these tasks on Replica.


\subsubsection{Open-Vocabulary Segmentation}
\label{sec:ovseg}


\textbf{Protocol.} We choose to evaluate OVSeg following a paradigm used in \cite{shi2024language} where we successively evaluate the segmentation performance of individual queries, which is a relevant paradigm for robotics applications. We also evaluate OVSeg performance on a simultaneous multi-class paradigm (used in \cite{engelmann2024opennerf}) in the supplementary materials.  For this experiment, we separate the evaluation in two and evaluate both the semantic \textit{classes} present in each Replica scene and the \textit{concepts} introduced in our extended dataset. \\
\steve{\textbf{Evaluated methods.} We evaluate the impact of plugging DiSCO (with 1 prototype) into both LeRF and OpenNeRF.}\\
\textbf{Results.}
We present quantitative outcomes in \autoref{tab:open-seg}, first analyzing results for \textit{classes}, followed by \textit{concepts}. Adding DiSCO improves segmentation performance across both LeRF and OpenNeRF feature fields. For \textit{class}-level segmentation, LeRF benefits from a +3.63 increase in mIoU and +3.40 in mAcc, demonstrating that DiSCO refines segmentation boundaries and mitigates feature noise. OpenNeRF, which already provides strong segmentation, also sees slight improvements in mIoU and mAcc. For \textit{concept}-level segmentation, DiSCO leads to even greater improvements, particularly in mAcc, with a +8.94 and +15.48 gain for LeRF and OpenNeRF respectively, highlighting its ability to complete sparse relevancy heatmaps (as vague queries usually result in globally low relevancy across a scene). Overall, DiSCO mitigates common open-vocabulary segmentation issues by reducing relevancy spilling for highly responsive queries, preventing over-segmentation, and filling relevancy holes for more complex queries, ensuring a more complete representation of semantic concepts.

\begin{table}[]
\centering
\scalebox{0.81}{
    \begin{tabular}{c cc cc }
    \hline
        \multirow{2}{*}{Method} &
          \multicolumn{2}{c}{\textit{Classes}} &
          \multicolumn{2}{c}{\textit{Concepts}} 
           \\ \cline{2-5}
        & mIoU $\uparrow$  & mAcc $\uparrow$ & mIoU $\uparrow$ & mAcc $\uparrow$ \\ 
        \hline
    
    LeRF~\cite{kerr2023lerf}     &  8.79  &    84.53    & 10.42    &  37.79     \\ 
    LeRF + DiSCO-3D  &   \textbf{12.42}   &  \textbf{87.93}    &  \textbf{15.78}  &   \textbf{46.73}     \\ \hline 
    OpenNeRF~\cite{engelmann2024opennerf} &  21.60    &  91.87  & 15.59 & 42.69 \\ 
    OpenNeRF + DiSCO-3D  & \textbf{21.87}    & \textbf{92.66}     &  \textbf{16.69}  & \textbf{58.17} \\ \hline
    \end{tabular}
}
\caption{\textbf{DiSCO-3D Quantitative Evaluation for OV-Seg.}}
\label{tab:open-seg}
\end{table}

\subsubsection{Unsupervised Semantic Segmentation} 
\label{sec:uss_exp}

\textbf{Protocol.} For this experiment, the evaluated methods are requested to segment the scene into $N$ semantic classes. We follow usual USS evaluation pipelines and use Hungarian matching to link the semantic classes to ground-truth classes. Since Replica scenes contain different sets of objects, we predict for every method $N=10$ semantic classes and compare our results to the top-10 classes of each scene. CLIP prototypes are not used in this setting. \\ 
\textbf{Evaluated methods.} Since no NeRF-based unsupervised semantic segmentation (USS) methods exist for direct comparison with DiSCO-3D, we construct a hand-crafted NeRF baseline where we extract DINO features per point of the evaluated point cloud using the feature field and apply K-Means clustering. Additionally, we evaluate two representative USS methods: the 2D method SmooSeg~\cite{lan2024smooseg} and the 3D point-cloud method GrowSP~\cite{zhang2023growsp}. Since SmooSeg only produces 2D segmentations, we recover a 3D segmentation by training a Semantic-NeRF~\cite{zhi2021inplace} on its outputs. \\ 
\textbf{Results.} Quantitative results can be found in \autoref{tab:uss}. 
DiSCO-3D achieves the best results across all evaluations. Firstly, 2D USS methods such as SmooSeg do not assure multi-view consistency meaning that one object seen from different viewpoints will have different semantic predictions. This impairs the 3D segmentation performances when training the Semantic-NeRF as NeRF predictions are agnostic to the viewpoint by design. Regarding GrowSP, although it succeeds in performing accurate segmentation, the global performances are lower, probably due to the input data modalities, as the discrete nature of point clouds may limit their expressiveness compared to the continuous representations of NeRF. Finally, K-Means on the feature field yields slightly lower but comparable results. However, it remains restricted to USS, as it cannot incorporate open-vocabulary queries for OV-SD. In contrast, DiSCO-3D effectively handles both OV-SD and its edge cases, OV-Seg and USS, demonstrating its versatility.
\begin{table}[]
\centering
\scalebox{0.90}{
\begin{tabular}{ll ll}
\hline
Paradigm    &  Method  & mIoU $\uparrow$  & mAcc $\uparrow$   \\ \hline
2D + NeRF & SmooSeg~\cite{lan2024smooseg}   &  10.49  &   31.07         \\ 

Point-Cloud  &  GrowSP~\cite{zhang2023growsp}   &  21.62   &  34.24 \\ 
\doriand{NeRF} & \doriand{K-Means} &    \doriand{27.00}    &  \doriand{50.14}     \\
NeRF &     DiSCO-3D      &  \textbf{27.47}    &  \textbf{ 51.99  }        \\ \hline
\end{tabular}
}
\caption{\textbf{DiSCO-3D Quantitative Evaluation for USS.} GrowSP uses features obtained from SparseConv while every other baselines uses DINO as input.}

\label{tab:uss}
\end{table}

\section{Conclusion}
\label{sec:ccl}

In this paper, we introduced the problem of 3D Open-Vocabulary Sub-Concept Discovery and presented a solution tailored to 3D Neural Field representations. Our approach combines an Unsupervised Semantic Segmentation module—the first designed for NeRF—with partial supervision from Open-Vocabulary Segmentation. 
While developed for Neural Fields, this method could theoretically be extended to other representations, such as 2D images, 3D point clouds, or 3D Gaussian Splatting. We believe that this new OV-SD challenge holds significant potential for practical applications and hope that this paper will inspire future research in the field.

\section*{Acknowledgements} This publication was made possible by the use of the CEA List FactoryIA supercomputer, financially supported by the Ile-de-France Regional Council.

{
    \small
    \bibliographystyle{ieeenat_fullname}
    \bibliography{main}

\begin{thebibliography}{41}
\providecommand{\natexlab}[1]{#1}
\providecommand{\url}[1]{\texttt{#1}}
\expandafter\ifx\csname urlstyle\endcsname\relax
  \providecommand{\doi}[1]{doi: #1}\else
  \providecommand{\doi}{doi: \begingroup \urlstyle{rm}\Url}\fi

\bibitem[Asgari~Taghanaki et~al.(2021)Asgari~Taghanaki, Abhishek, Cohen,
  Cohen-Adad, and Hamarneh]{asgari2021deep}
Saeid Asgari~Taghanaki, Kumar Abhishek, Joseph~Paul Cohen, Julien Cohen-Adad,
  and Ghassan Hamarneh.
\newblock Deep semantic segmentation of natural and medical images: a review.
\newblock \emph{Artificial Intelligence Review}, 54:\penalty0 137--178, 2021.

\bibitem[Barron et~al.(2022)Barron, Mildenhall, Verbin, Srinivasan, and
  Hedman]{barron2022mip}
Jonathan~T Barron, Ben Mildenhall, Dor Verbin, Pratul~P Srinivasan, and Peter
  Hedman.
\newblock Mip-nerf 360: Unbounded anti-aliased neural radiance fields.
\newblock In \emph{Proceedings of the IEEE/CVF Conference on Computer Vision
  and Pattern Recognition}, pages 5470--5479, 2022.

\bibitem[Caron et~al.(2021)Caron, Touvron, Misra, J{\'e}gou, Mairal,
  Bojanowski, and Joulin]{caron2021emerging}
Mathilde Caron, Hugo Touvron, Ishan Misra, Herv{\'e} J{\'e}gou, Julien Mairal,
  Piotr Bojanowski, and Armand Joulin.
\newblock Emerging properties in self-supervised vision transformers.
\newblock In \emph{Proceedings of the IEEE/CVF international conference on
  computer vision}, pages 9650--9660, 2021.

\bibitem[Chen et~al.(2023)Chen, Tang, Wan, Wang, and Zeng]{chen2023interactive}
Xiaokang Chen, Jiaxiang Tang, Diwen Wan, Jingbo Wang, and Gang Zeng.
\newblock Interactive segment anything nerf with feature imitation.
\newblock \emph{arXiv preprint arXiv:2305.16233}, 2023.

\bibitem[Eftekhar et~al.(2024)Eftekhar, Zeng, Duan, Farhadi, Kembhavi, and
  Krishna]{eftekhar2023selective}
Ainaz Eftekhar, Kuo-Hao Zeng, Jiafei Duan, Ali Farhadi, Ani Kembhavi, and
  Ranjay Krishna.
\newblock Selective visual representations improve convergence and
  generalization for embodied ai.
\newblock In \emph{ICLR}, 2024.

\bibitem[Engelmann et~al.(2024)Engelmann, Manhardt, Niemeyer, Tateno,
  Pollefeys, and Tombari]{engelmann2024opennerf}
Francis Engelmann, Fabian Manhardt, Michael Niemeyer, Keisuke Tateno, Marc
  Pollefeys, and Federico Tombari.
\newblock Opennerf: Open set 3d neural scene segmentation with pixel-wise
  features and rendered novel views.
\newblock \emph{arXiv preprint arXiv:2404.03650}, 2024.

\bibitem[Feng et~al.(2020)Feng, Haase-Sch{\"u}tz, Rosenbaum, Hertlein, Glaeser,
  Timm, Wiesbeck, and Dietmayer]{feng2020deep}
Di Feng, Christian Haase-Sch{\"u}tz, Lars Rosenbaum, Heinz Hertlein, Claudius
  Glaeser, Fabian Timm, Werner Wiesbeck, and Klaus Dietmayer.
\newblock Deep multi-modal object detection and semantic segmentation for
  autonomous driving: Datasets, methods, and challenges.
\newblock \emph{IEEE Transactions on Intelligent Transportation Systems},
  22\penalty0 (3):\penalty0 1341--1360, 2020.

\bibitem[Ghiasi et~al.(2022)Ghiasi, Gu, Cui, and Lin]{ghiasi2022scaling}
Golnaz Ghiasi, Xiuye Gu, Yin Cui, and Tsung-Yi Lin.
\newblock Scaling open-vocabulary image segmentation with image-level labels.
\newblock In \emph{European Conference on Computer Vision}, pages 540--557.
  Springer, 2022.

\bibitem[Hamilton et~al.(2022)Hamilton, Zhang, Hariharan, Snavely, and
  Freeman]{hamilton2022unsupervised}
Mark Hamilton, Zhoutong Zhang, Bharath Hariharan, Noah Snavely, and William~T
  Freeman.
\newblock Unsupervised semantic segmentation by distilling feature
  correspondences.
\newblock \emph{arXiv preprint arXiv:2203.08414}, 2022.

\bibitem[Hurtado and Valada(2022)]{hurtado2022semantic}
Juana~Valeria Hurtado and Abhinav Valada.
\newblock Semantic scene segmentation for robotics.
\newblock In \emph{Deep learning for robot perception and cognition}, pages
  279--311. Elsevier, 2022.

\bibitem[Kerbl et~al.(2023)Kerbl, Kopanas, Leimk{\"u}hler, and
  Drettakis]{kerbl3Dgaussians}
Bernhard Kerbl, Georgios Kopanas, Thomas Leimk{\"u}hler, and George Drettakis.
\newblock 3d gaussian splatting for real-time radiance field rendering.
\newblock \emph{ACM Transactions on Graphics}, 42\penalty0 (4), 2023.

\bibitem[Kerr et~al.(2023)Kerr, Kim, Goldberg, Kanazawa, and
  Tancik]{kerr2023lerf}
Justin* Kerr, Chung~Min* Kim, Ken Goldberg, Angjoo Kanazawa, and Matthew
  Tancik.
\newblock Lerf: Language embedded radiance fields.
\newblock In \emph{International Conference on Computer Vision (ICCV)}, 2023.

\bibitem[Kim et~al.(2024{\natexlab{a}})Kim, Han, Ju, and Hwang]{kim2024eagle}
Chanyoung Kim, Woojung Han, Dayun Ju, and Seong~Jae Hwang.
\newblock Eagle: Eigen aggregation learning for object-centric unsupervised
  semantic segmentation.
\newblock In \emph{Proceedings of the IEEE/CVF Conference on Computer Vision
  and Pattern Recognition}, pages 3523--3533, 2024{\natexlab{a}}.

\bibitem[Kim et~al.(2024{\natexlab{b}})Kim, Wu, Kerr, Goldberg, Tancik, and
  Kanazawa]{kim2024garfield}
Chung~Min Kim, Mingxuan Wu, Justin Kerr, Ken Goldberg, Matthew Tancik, and
  Angjoo Kanazawa.
\newblock Garfield: Group anything with radiance fields.
\newblock In \emph{Proceedings of the IEEE/CVF Conference on Computer Vision
  and Pattern Recognition}, pages 21530--21539, 2024{\natexlab{b}}.

\bibitem[Kirillov et~al.(2019)Kirillov, He, Girshick, Rother, and
  Doll{\'a}r]{kirillov2019panoptic}
Alexander Kirillov, Kaiming He, Ross Girshick, Carsten Rother, and Piotr
  Doll{\'a}r.
\newblock Panoptic segmentation.
\newblock In \emph{Proceedings of the IEEE/CVF conference on computer vision
  and pattern recognition}, pages 9404--9413, 2019.

\bibitem[Kirillov et~al.(2023)Kirillov, Mintun, Ravi, Mao, Rolland, Gustafson,
  Xiao, Whitehead, Berg, Lo, et~al.]{kirillov2023segment}
Alexander Kirillov, Eric Mintun, Nikhila Ravi, Hanzi Mao, Chloe Rolland, Laura
  Gustafson, Tete Xiao, Spencer Whitehead, Alexander~C Berg, Wan-Yen Lo, et~al.
\newblock Segment anything.
\newblock In \emph{Proceedings of the IEEE/CVF International Conference on
  Computer Vision}, pages 4015--4026, 2023.

\bibitem[Kobayashi et~al.(2022)Kobayashi, Matsumoto, and
  Sitzmann]{kobayashi2022decomposing}
Sosuke Kobayashi, Eiichi Matsumoto, and Vincent Sitzmann.
\newblock Decomposing nerf for editing via feature field distillation.
\newblock \emph{Advances in Neural Information Processing Systems},
  35:\penalty0 23311--23330, 2022.

\bibitem[Lan et~al.(2024)Lan, Wang, Ke, Xu, Feng, and Zhang]{lan2024smooseg}
Mengcheng Lan, Xinjiang Wang, Yiping Ke, Jiaxing Xu, Litong Feng, and Wayne
  Zhang.
\newblock Smooseg: smoothness prior for unsupervised semantic segmentation.
\newblock \emph{Advances in Neural Information Processing Systems}, 36, 2024.

\bibitem[Li et~al.(2022)Li, Weinberger, Belongie, Koltun, and
  Ranftl]{li2022language}
Boyi Li, Kilian~Q Weinberger, Serge Belongie, Vladlen Koltun, and Ren{\'e}
  Ranftl.
\newblock Language-driven semantic segmentation.
\newblock \emph{ICLR}, 2022.

\bibitem[Li et~al.(2023)Li, Wang, Cheng, Yu, Zhao, Song, Liu, Yuan, and
  Chen]{li2023acseg}
Kehan Li, Zhennan Wang, Zesen Cheng, Runyi Yu, Yian Zhao, Guoli Song, Chang
  Liu, Li Yuan, and Jie Chen.
\newblock Acseg: Adaptive conceptualization for unsupervised semantic
  segmentation.
\newblock In \emph{Proceedings of the IEEE/CVF conference on computer vision
  and pattern recognition}, pages 7162--7172, 2023.

\bibitem[Ling et~al.(2024)Ling, Sheng, Tu, Zhao, Xin, Wan, Yu, Guo, Yu, Lu,
  et~al.]{ling2024dl3dv}
Lu Ling, Yichen Sheng, Zhi Tu, Wentian Zhao, Cheng Xin, Kun Wan, Lantao Yu,
  Qianyu Guo, Zixun Yu, Yawen Lu, et~al.
\newblock Dl3dv-10k: A large-scale scene dataset for deep learning-based 3d
  vision.
\newblock In \emph{Proceedings of the IEEE/CVF Conference on Computer Vision
  and Pattern Recognition}, pages 22160--22169, 2024.

\bibitem[Liu et~al.(2024{\natexlab{a}})Liu, Yu, Breckon, and
  Shum]{liu2024u3ds3}
Jiaxu Liu, Zhengdi Yu, Toby~P Breckon, and Hubert~PH Shum.
\newblock U3ds3: Unsupervised 3d semantic scene segmentation.
\newblock In \emph{Proceedings of the IEEE/CVF Winter Conference on
  Applications of Computer Vision}, pages 3759--3768, 2024{\natexlab{a}}.

\bibitem[Liu et~al.(2023)Liu, Zhan, Zhang, Xu, Yu, El~Saddik, Theobalt, Xing,
  and Lu]{liu2023weakly}
Kunhao Liu, Fangneng Zhan, Jiahui Zhang, Muyu Xu, Yingchen Yu, Abdulmotaleb
  El~Saddik, Christian Theobalt, Eric Xing, and Shijian Lu.
\newblock Weakly supervised 3d open-vocabulary segmentation.
\newblock \emph{Advances in Neural Information Processing Systems},
  36:\penalty0 53433--53456, 2023.

\bibitem[Liu et~al.(2024{\natexlab{b}})Liu, Hu, Tang, and Tai]{liu2024sanerf}
Yichen Liu, Benran Hu, Chi-Keung Tang, and Yu-Wing Tai.
\newblock Sanerf-hq: Segment anything for nerf in high quality.
\newblock In \emph{Proceedings of the IEEE/CVF Conference on Computer Vision
  and Pattern Recognition}, pages 3216--3226, 2024{\natexlab{b}}.

\bibitem[Mildenhall et~al.(2021)Mildenhall, Srinivasan, Tancik, Barron,
  Ramamoorthi, and Ng]{mildenhall2021nerf}
Ben Mildenhall, Pratul~P Srinivasan, Matthew Tancik, Jonathan~T Barron, Ravi
  Ramamoorthi, and Ren Ng.
\newblock Nerf: Representing scenes as neural radiance fields for view
  synthesis.
\newblock \emph{Communications of the ACM}, 65\penalty0 (1):\penalty0 99--106,
  2021.

\bibitem[Mo et~al.(2022)Mo, Wu, Yang, Liu, and Liao]{mo2022review}
Yujian Mo, Yan Wu, Xinneng Yang, Feilin Liu, and Yujun Liao.
\newblock Review the state-of-the-art technologies of semantic segmentation
  based on deep learning.
\newblock \emph{Neurocomputing}, 493:\penalty0 626--646, 2022.

\bibitem[M\"uller et~al.(2022)M\"uller, Evans, Schied, and
  Keller]{mueller2022instant}
Thomas M\"uller, Alex Evans, Christoph Schied, and Alexander Keller.
\newblock Instant neural graphics primitives with a multiresolution hash
  encoding.
\newblock \emph{ACM Trans. Graph.}, 41\penalty0 (4):\penalty0 102:1--102:15,
  2022.

\bibitem[Oquab et~al.(2023)Oquab, Darcet, Moutakanni, Vo, Szafraniec, Khalidov,
  Fernandez, Haziza, Massa, El-Nouby, et~al.]{oquab2023dinov2}
Maxime Oquab, Timoth{\'e}e Darcet, Th{\'e}o Moutakanni, Huy Vo, Marc
  Szafraniec, Vasil Khalidov, Pierre Fernandez, Daniel Haziza, Francisco Massa,
  Alaaeldin El-Nouby, et~al.
\newblock Dinov2: Learning robust visual features without supervision.
\newblock \emph{arXiv preprint arXiv:2304.07193}, 2023.

\bibitem[Peng et~al.(2023)Peng, Genova, Jiang, Tagliasacchi, Pollefeys,
  Funkhouser, et~al.]{peng2023openscene}
Songyou Peng, Kyle Genova, Chiyu Jiang, Andrea Tagliasacchi, Marc Pollefeys,
  Thomas Funkhouser, et~al.
\newblock Openscene: 3d scene understanding with open vocabularies.
\newblock In \emph{Proceedings of the IEEE/CVF conference on computer vision
  and pattern recognition}, pages 815--824, 2023.

\bibitem[Radford et~al.(2021)Radford, Kim, Hallacy, Ramesh, Goh, Agarwal,
  Sastry, Askell, Mishkin, Clark, et~al.]{radford2021learning}
Alec Radford, Jong~Wook Kim, Chris Hallacy, Aditya Ramesh, Gabriel Goh,
  Sandhini Agarwal, Girish Sastry, Amanda Askell, Pamela Mishkin, Jack Clark,
  et~al.
\newblock Learning transferable visual models from natural language
  supervision.
\newblock In \emph{International conference on machine learning}, pages
  8748--8763. PMLR, 2021.

\bibitem[Shen et~al.(2023)Shen, Yang, Yu, Wong, Kaelbling, and
  Isola]{shen2023distilled}
William Shen, Ge Yang, Alan Yu, Jansen Wong, Leslie~Pack Kaelbling, and Phillip
  Isola.
\newblock Distilled feature fields enable few-shot language-guided
  manipulation.
\newblock \emph{arXiv preprint arXiv:2308.07931}, 2023.

\bibitem[Shi et~al.(2024)Shi, Wang, Duan, and Guan]{shi2024language}
Jin-Chuan Shi, Miao Wang, Hao-Bin Duan, and Shao-Hua Guan.
\newblock Language embedded 3d gaussians for open-vocabulary scene
  understanding.
\newblock In \emph{Proceedings of the IEEE/CVF Conference on Computer Vision
  and Pattern Recognition}, pages 5333--5343, 2024.

\bibitem[Straub et~al.(2019)Straub, Whelan, Ma, Chen, Wijmans, Green, Engel,
  Mur-Artal, Ren, Verma, Clarkson, Yan, Budge, Yan, Pan, Yon, Zou, Leon,
  Carter, Briales, Gillingham, Mueggler, Pesqueira, Savva, Batra, Strasdat,
  Nardi, Goesele, Lovegrove, and Newcombe]{replica19arxiv}
Julian Straub, Thomas Whelan, Lingni Ma, Yufan Chen, Erik Wijmans, Simon Green,
  Jakob~J. Engel, Raul Mur-Artal, Carl Ren, Shobhit Verma, Anton Clarkson,
  Mingfei Yan, Brian Budge, Yajie Yan, Xiaqing Pan, June Yon, Yuyang Zou,
  Kimberly Leon, Nigel Carter, Jesus Briales, Tyler Gillingham, Elias Mueggler,
  Luis Pesqueira, Manolis Savva, Dhruv Batra, Hauke~M. Strasdat, Renzo~De
  Nardi, Michael Goesele, Steven Lovegrove, and Richard Newcombe.
\newblock The {R}eplica dataset: A digital replica of indoor spaces.
\newblock \emph{arXiv preprint arXiv:1906.05797}, 2019.

\bibitem[Tancik et~al.(2023)Tancik, Weber, Ng, Li, Yi, Wang, Kristoffersen,
  Austin, Salahi, Ahuja, et~al.]{tancik2023nerfstudio}
Matthew Tancik, Ethan Weber, Evonne Ng, Ruilong Li, Brent Yi, Terrance Wang,
  Alexander Kristoffersen, Jake Austin, Kamyar Salahi, Abhik Ahuja, et~al.
\newblock Nerfstudio: A modular framework for neural radiance field
  development.
\newblock In \emph{ACM SIGGRAPH 2023 Conference Proceedings}, pages 1--12,
  2023.

\bibitem[Tschernezki et~al.(2022)Tschernezki, Laina, Larlus, and
  Vedaldi]{tschernezki2022neural}
Vadim Tschernezki, Iro Laina, Diane Larlus, and Andrea Vedaldi.
\newblock Neural feature fusion fields: 3d distillation of self-supervised 2d
  image representations.
\newblock In \emph{2022 International Conference on 3D Vision (3DV)}, pages
  443--453. IEEE, 2022.

\bibitem[Wysocza{\'n}ska et~al.(2024)Wysocza{\'n}ska, Sim{\'e}oni,
  Ramamonjisoa, Bursuc, Trzci{\'n}ski, and P{\'e}rez]{wysoczanska2024clipdino}
Monika Wysocza{\'n}ska, Oriane Sim{\'e}oni, Micha{\"e}l Ramamonjisoa, Andrei
  Bursuc, Tomasz Trzci{\'n}ski, and Patrick P{\'e}rez.
\newblock Clip-dinoiser: Teaching clip a few dino tricks for open-vocabulary
  semantic segmentation.
\newblock \emph{ECCV}, 2024.

\bibitem[Zhang et~al.(2023)Zhang, Yang, Wang, and Li]{zhang2023growsp}
Zihui Zhang, Bo Yang, Bing Wang, and Bo Li.
\newblock Growsp: Unsupervised semantic segmentation of 3d point clouds.
\newblock In \emph{Proceedings of the IEEE/CVF Conference on Computer Vision
  and Pattern Recognition}, pages 17619--17629, 2023.

\bibitem[Zheng~Ding(2023)]{ding2023maskclip}
Zhuowen~Tu Zheng~Ding, Jieke~Wang.
\newblock Open-vocabulary universal image segmentation with maskclip.
\newblock In \emph{International Conference on Machine Learning}, 2023.

\bibitem[Zhi et~al.(2021)Zhi, Laidlow, Leutenegger, and
  Davison]{zhi2021inplace}
Shuaifeng Zhi, Tristan Laidlow, Stefan Leutenegger, and Andrew Davison.
\newblock In-place scene labelling and understanding with implicit scene
  representation.
\newblock In \emph{Proceedings of the International Conference on Computer
  Vision (ICCV)}, 2021.

\bibitem[Zhou et~al.(2022)Zhou, Ren, Xu, Liu, and Zhou]{zhou2022supervised}
Yuguo Zhou, Yanbo Ren, Erya Xu, Shiliang Liu, and Lijian Zhou.
\newblock Supervised semantic segmentation based on deep learning: a survey.
\newblock \emph{Multimedia Tools and Applications}, 81\penalty0 (20):\penalty0
  29283--29304, 2022.

\bibitem[Zou et~al.(2023)Zou, Dou, Yang, Gan, Li, Li, Dai, Behl, Wang, Yuan,
  et~al.]{zou2023generalized}
Xueyan Zou, Zi-Yi Dou, Jianwei Yang, Zhe Gan, Linjie Li, Chunyuan Li, Xiyang
  Dai, Harkirat Behl, Jianfeng Wang, Lu Yuan, et~al.
\newblock Generalized decoding for pixel, image, and language.
\newblock In \emph{Proceedings of the IEEE/CVF Conference on Computer Vision
  and Pattern Recognition}, pages 15116--15127, 2023.

\end{thebibliography}
}

\clearpage
\setcounter{page}{1}
\maketitlesupplementary


\newcommand{\cmark}{\ding{51}}%
\newcommand{\xmark}{\ding{55}}%

\interfootnotelinepenalty=10000


\section{DiSCO-3D}
\label{sec:supmat}

\subsection{Additional Architecture Details}

\begin{figure}[t]
    \includegraphics[width=0.9\columnwidth]{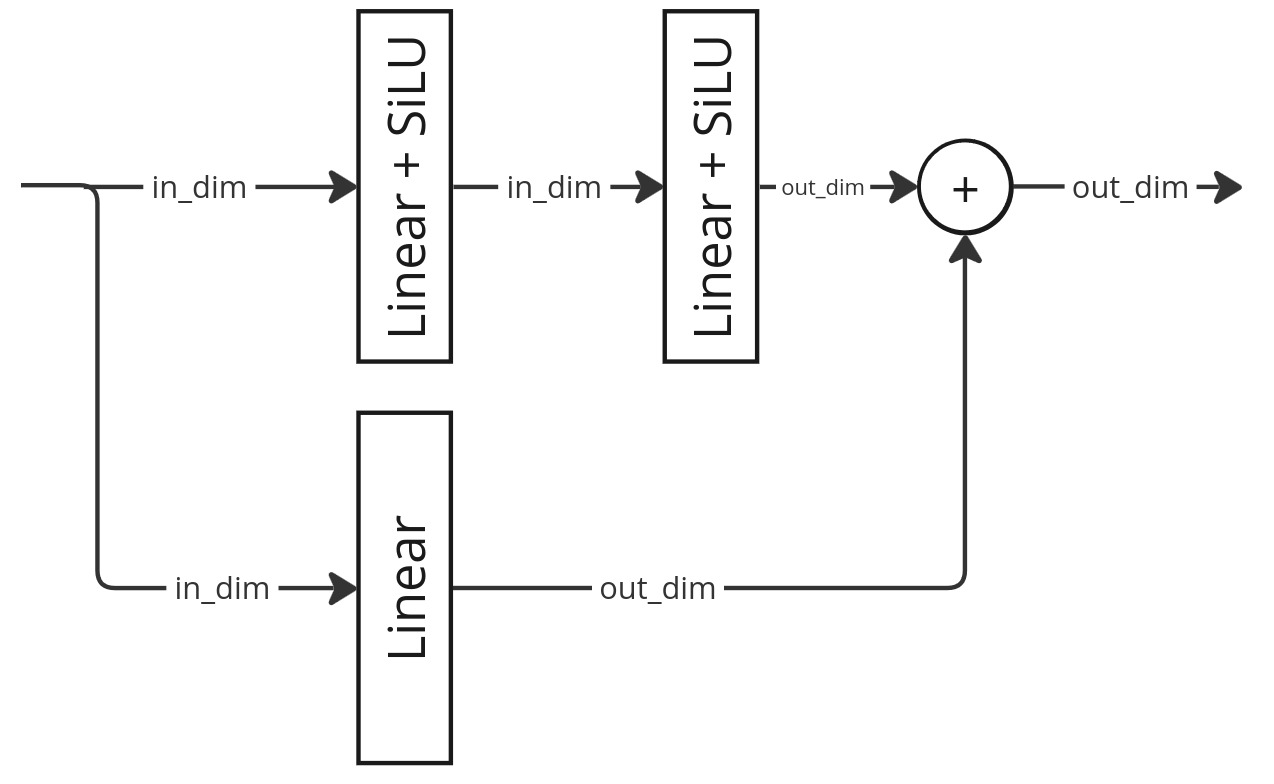}
    \caption{\textbf{Projector Architecture.}}
    \label{fig:projector}
\end{figure}

Some architecture details and minor contributions have been overlooked in the main paper that we want to cover here. \\
\textbf{Projector Architecture.} Although some USS methods implement a simple linear MLP projector \cite{hamilton2022unsupervised}, we follow SmooSeg \cite{lan2024smooseg} and decide to use a slightly more complex architecture depicted in \autoref{fig:projector}. It combines a non-linear MLP (with SiLU activations) with a linear layer serving as a residual connection. It is however to be noted that the impact (in both quality and time) is minimal, as evaluated in the following ablative experiments of the supplementary material. \\
\textbf{Filtering Uncertain Samples.} The prototypes of DiSCO-3D are updated each epoch via an EMA with a two-fold weighted average on both the density weights of the batch's samples and the prediction confidence (corresponding to the prediction probability of the class). In practise, we decide to further regularize this EMA by filtering out of the update process samples with very low weights (both density and confidence weights). For each sample $k$ classified as $i$, if $D_{k,i} < 0.2$ or $w_k < 0.2$, the related feature $f^{proj}_k$ will not participate in the update process.  \\

\subsection{LeRF Multi-scale CLIP Pyramid}
Because CLIP outputs an embedding per image, rather than pixel-wise embeddings, it is not trivial to encode a scene as a CLIP feature field. While some methods work on adapting CLIP to pixel-wise embeddings \cite{li2022language,ghiasi2022scaling,ding2023maskclip,wysoczanska2024clipdino} (for instance, OpenSeg, which is a feature field used in our paper proposes a CLIP model adapted for dense tasks such as segmentation), LeRF proposes to pass image patches of different sizes into CLIP to produce a multi-scale pyramid used as supervision material. Regarding the LeRF model in itself, a scale parameter is added as input to the feature decoder and the training is done by randomly sampling scales across the pyramid for each sampled ray and retrieving the associated CLIP feature. During inference, the relevancy related to a query is computed for a pre-defined number of different scales and we display the relevancy heatmap of the scale resulting in maximum global relevancy, as done in \autoref{fig:limit1}, \autoref{fig:limit2} and \autoref{fig:spill}.  \\

In order to accommodate DiSCO-3D to this multi-scale pyramid when plugging into LeRF, several small modifications are made on the CLIP branch (no changes on the DINO branch because DINO produces pixel-wise embeddings). For each sample at each epoch, we decipher the associated CLIP embedding to be used for the computation of $\calL_{irr}$ and $\calP^{CLIP}$ by choosing the scale which outputs the maximum similarity to the user's query. This computation is performed by evaluating the similarity on a discrete number of scales, as done in LeRF inference (except we use the per-sample maximum similarity rather than per-image).

Note that when we use an empty query (i.e. when doing unsupervised semantic segmentation), CLIP prototypes can be computed using random scales for each samples. Multi-scales prototypes (which stores an average CLIP embedding per scale) has been tested, with a minimal increase of performance for an important increase of compute duration.




\subsection{Notion of Confidence in DiSCO-3D}

\begin{figure}[t]
    \includegraphics[width=\columnwidth]{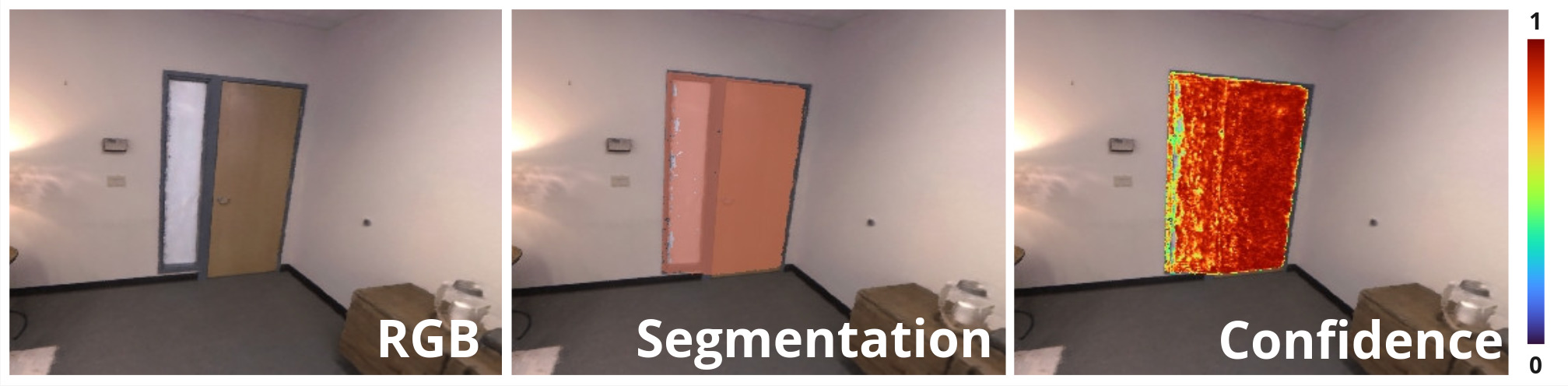}
    \caption{\textbf{Segmentation Confidence of DiSCO-3D.} The query is "door". }
    \label{fig:confidence}
\end{figure}

Although DiSCO-3D performs open-vocabulary segmentation (with hard class assignment rather than relevancy computation as in LeRF and OpenNeRF), confidence scores can be obtained by using the probability distributions $D$ after the softmax operation. These scores define how similar each sample's post-projection feature is to its associated prototype compared to the other prototypes. \autoref{fig:confidence} illustrates a confidence heatmap for the query "door". The predictions are globally confident, which is normal as DiSCO-3D encourages high confidence by design (especially with the $\beta$ scheduling defined in~\autoref{sec:USS}). However, we can notice less confidence on the door edges, and especially on the narrow window at the left of the door, which is rather coherent as it could arguably not be considered as part of the door.


\subsection{Limitations and Failure Cases}

\begin{figure}[t]
    \includegraphics[width=\columnwidth]{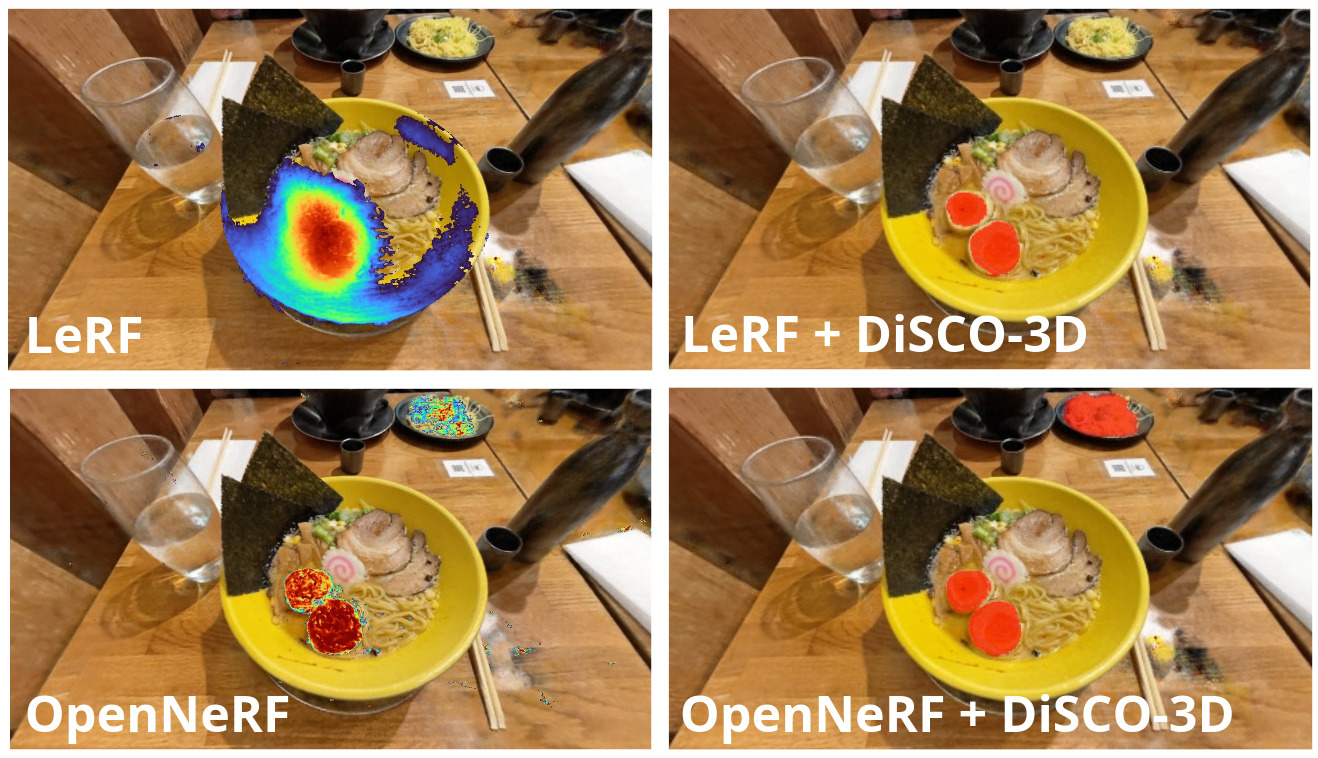}
    \caption{\textbf{Limitation $\#1$.} By querying "Eggs", the LeRF and OpenNeRF baselines makes different prediction, both regarding the responding objects and their precision. While DiSCO-3D can "repair" segmentation imprecision via the DINO features, it is dependent of the open-vocabulary expressivity making the OpenNeRF+DiSCO also segment the background dish even though it does not seem to contain eggs.}
    \label{fig:limit1}
\end{figure}

\begin{figure}[t]
    \includegraphics[width=\columnwidth]{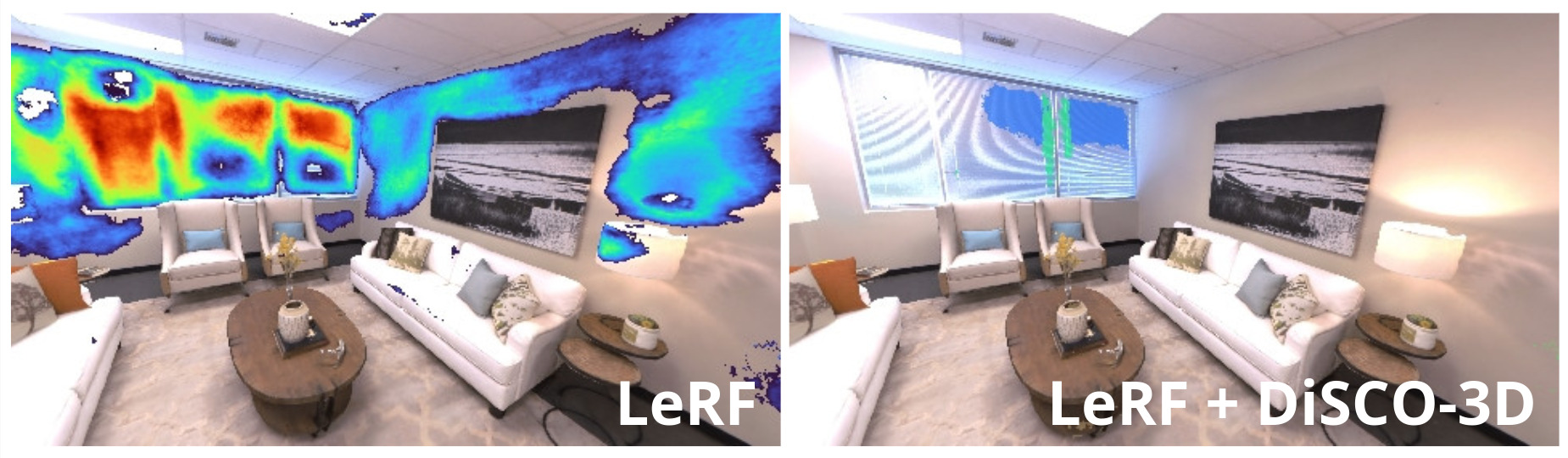}
    \caption{\textbf{Limitation $\#2$.} We query "Art", which is incorrectly detected in LeRF, resulting in an irrelevant segmentation of parts of the windows rather than the painting.}
    \label{fig:limit2}
\end{figure}

Here, we discuss and illustrate a number of limitations inherent to our method. \\
\textbf{Feature Field Quality Dependent.} First of all, we mentioned the dependency of our method to the pre-trained feature field performance. Since this field provides input features for both segmentation and open-vocabulary queries, inaccuracies can negatively impact the results, as illustrated in \autoref{fig:limit1} and \autoref{fig:limit2}. Regarding the open-vocabulary field, errors are common due to the limited quality of 2D open-vocabulary models and inaccuracies in NeRF’s 3D projection, often caused by imprecise camera poses. These errors can lead to unexpected query results—either an excessive number of objects being labeled as relevant (e.g., the "Eggs" example in \autoref{fig:limit1}) or a failure to correctly interpret some queries, especially when they regard abstract concepts, preventing DiSCO-3D from segmenting the intended sub-concepts. An example of the latter issue is shown in \autoref{fig:limit2}, where the query "Art" fails to recognize the painting while incorrectly identifying seemingly random parts of the scene. This confusion propagates through the model, leading to incorrect segmentations. The projector feature field (e.g. DINO) can also suffer some issues which can have an impact on DiSCO's  performances. Depending on the used encoder, some models like DINO tend to produce features which describes the scene at object parts-level rather than object-level. This can lead sometimes to over-segmentation of sub-concepts. Although this may be useful in certain applications (eg. object decomposition), this phenomenom is not wanted in OV-SD and this is why we proposed the $\mathcal{L}_{proto}$ to reduce this over-segmentation. \\

\begin{figure*}[t]
    \includegraphics[width=\textwidth]{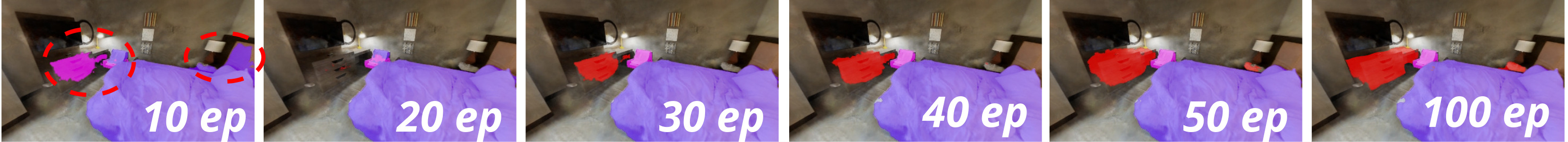}
    \caption{\textbf{Optimization Timelapse.} In average, one epoch takes \textbf{22ms}, resulting in a training of 200 epochs in \textbf{$\sim4$s}. The query is "furniture".}
    \label{fig:timelapse}
\end{figure*}

Although these issues originate from the input feature fields not introduced by our method, we can derive a few perspectives to improve the performances, which can be ordered in two classes. First, we can simply improve the quality of the feature fields, notably by using newer better image encoders, as discussed in the next subsection. On the other hand, we can work on the robustness of DiSCO-3D to mitigate the described issues. Although major failures caused by the input feature fields are hardly solvable, architecture improvements could be studied to incorporate more 3D geometry coherency in the segmentation process. \\
\textbf{Query-Specific Optimization.} Contrary to similarity-based open-vocabulary NeRF methods which only rely on a forward pass of their model to process a user query, DiSCO needs an optimization process of both the projector and the prototypes for each query to perform segmentation. However, we insist that the optimization is very fast, necessitating only very few and fast epochs to converge. Indeed, we typically achieve convergence in less than 100 epochs of approximately 20ms each, averaging a standard training of 2s. In~\autoref{fig:timelapse}, we display the evolution of the segmentation during the optimization process. While this per-query optimization limits for now true real-time processing, we believe the optimization to be fast enough for the method to be truly useful and applicable in real-life scenario. 

\begin{figure}[tb]
    \includegraphics[width=\columnwidth]{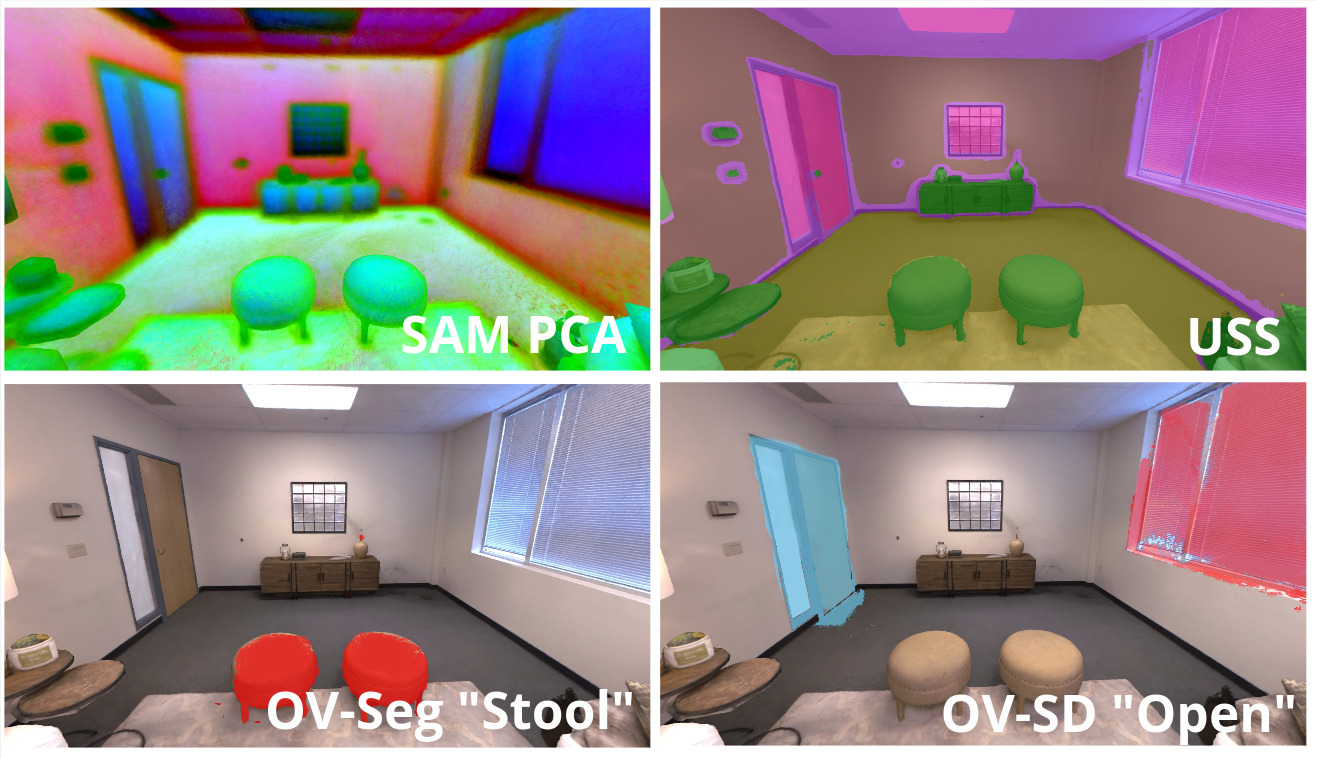}
    \caption{\textbf{SAM Feature Field.} We replace the DINO feature field in LeRF by a SAM feature field and demonstrate its capacity to perform USS, OV-Seg and OV-SD.}
    \label{fig:sam}
\end{figure}

\begin{figure}[tb]
    \includegraphics[width=\columnwidth]{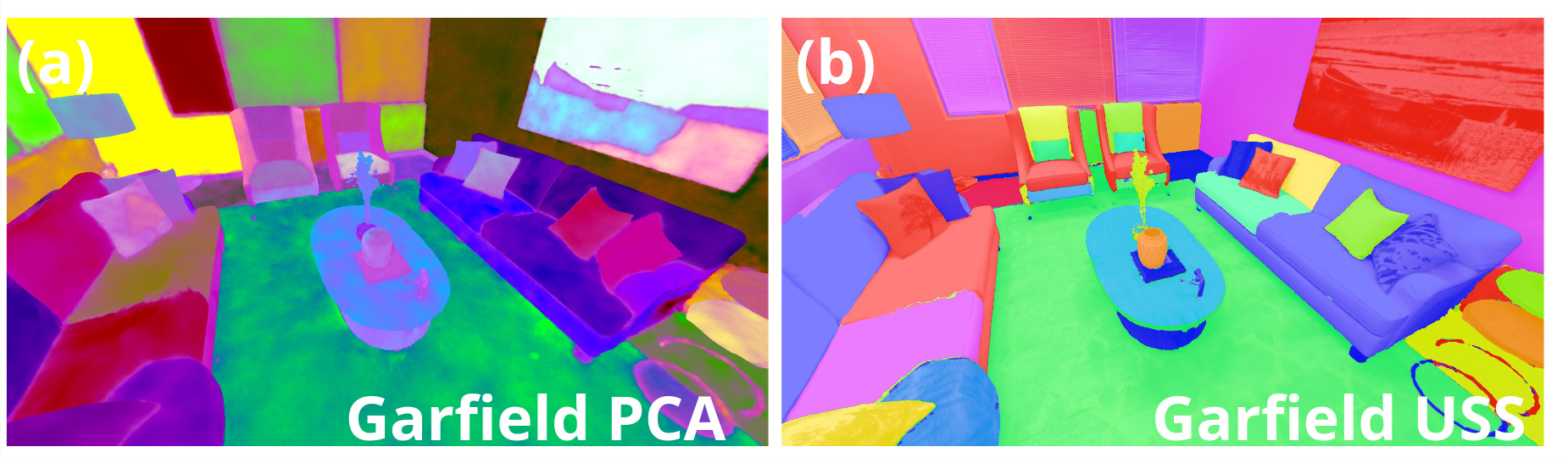}
    \caption{\textbf{Garfield Feature Field.} We use Garfield (SAM Masks outputs) as the segmentation field and perform USS. Note that Garfield being an instance feature field, it cannot be used as a replacement for DINO to perform OVSeg and OV-SD. USS also cannot be entirely considered as semantic segmentation.}
    \label{fig:garfield}
\end{figure}

\subsection{Extension to other feature fields}

We introduced in \autoref{sec:extensions} the possibility to use different feature fields, as long as we have a queriable feature field to serve as the query latent space and a spatially precise one to serve as input to the projector. In the main paper, the query feature space has been tested only with a multi-scale CLIP and the dense OpenSeg while the input to the projector has been respectively a DINO and OpenSeg feature field. \\

Regarding the segmentation feature field, there exists a large range of precise image encoders that can be injected into a feature field. For instance, \autoref{fig:sam} shows an example of DiSCO-3D applied on a modified LeRF where the DINO is replaced by the image encoder of SAM (without the decoder). Because SAM is also quite spatially precise (as shown in the PCA), it can successfully be used to perform any of the 3 proposed tasks (OV-SD, OVSeg and USS). In~\autoref{fig:garfield}, we display another example of segmentation feature field by using a Garfield feature field. Garfield is a method producing a multi-scale feature field using SAM segmentation masks and contrastive learning. However, for better understanding, we limit here the Garfield feature field to mono-scale segmentation. This results in extremely precise (but over-segmented) scene decomposition as shown in the PCA which can be used to perform unsupervised segmentation. However, it is important to note that SAM produces instance segmentation masks that are unaware of the semantics. Hence, they cannot be used to perform true USS, nor OV-Seg and OV-SD, but rather instance segmentation. Future works could focus on combining Garfield with previously introduced semantic fields to perform both semantic and instance segmentation at once. \\

Regarding the query feature field, we evaluated in the main paper two adaptations of the open-vocabulary model CLIP (LeRF and OpenNeRF). However, we could broaden the range of feature fields used for the query, using other open-vocabulary models for instance or change the modality of the query with other feature spaces (e.g. image queries with DINO encodings or user clicks with any feature, as shown in \autoref{fig:resu} with CLIP).


\section{Experiments}

\subsection{Hyperparameters}

In this section, we list the used hyperparameters for our different experiments (both quantitative and qualitative) of the article. \\
\textbf{Base Nerfacto Model Configuration.} We use most of the default Nerfstudio setup including with 16 hash grids and a dictionary size of $2^{19}$. For quantitative experiments with Replica's synthetic scenes, we use a feature size of 2 and bump it to 8 for more complex real scenes used in qualitative experiments. We also disable the camera optimizer and appearance embedding on Replica, as they overcomplexify the models for no real gain in segmentation performance. Finally, for all indoor scenes, we reduce the far plane to the scene's maximum dimension.  \\
\textbf{Pre-trained Feature Fields.} The configurations for the feature fields follow standard setups defined by LeRF. We use a set of hashgrids disjoint from the Nerfacto grids of 24 levels ($2^{19}$ dict size) with 8 feature size and resolutions ranging from 16 to 512. Following both LeRF and OpenNeRF, we use respectively an OpenCLIP base (ViT-B/16) and a CLIP large (ViT-L/14). The DINO used for LeRF is a ViT-S/8. \\
\textbf{DiSCO-3D Hyperparameters.} Unlike 2D USS methods that cluster DINO features, which are notoriously sensitive to hyperparameter tuning and prone to failures on diverse datasets, DiSCO-3D benefits from NeRF's scene-specificity, making it more robust. However, while DiSCO-3D has relatively few hyperparameters, certain parameters still require careful adjustments.  

\begin{itemize}
    \item \textbf{Number of Prototypes.} We showed in \autoref{sec:subconceptDiscovery} that the chosen number of relevant prototypes is not crucial as long as there are enough to describe each sub-concept. Regarding the number of irrelevant prototypes, we use three irrelevant prototypes in all experiments. However, this is not a sensitive hyperparameter, only requiring sufficient expressivity to encompass diverse irrelevant objects.
    \item \textbf{Projector.} The projector follows the introduced architecture and uses linear layers which both have as hidden dimension and output dimension the input dimension (ie. the feature dim). A dropout of probability $p=0.2$ is also applied on it.
    \item \textbf{$\beta$ Scheduling.} The $\beta$ hyperparameter and its linear scheduling configuration, which affect the sharpness of the probability distributions, also exhibit minimal impact across scenes as long as we keep a sound configuration. In the experiments, we use an initial value of 0.5, linearly decreasing to 0.1 over the training.
    \item \textbf{Thresholds.} The threshold for $\calL_{proj}$ has little impact and is fixed at 0.5, but $\calL_{irr}$'s threshold is more crucial and depends on the feature field. Indeed, OpenNeRF with its OpenSeg encoder generally outputs higher relevancy scores than LeRF with CLIP. To accommodate this difference, we use distinct thresholds: thresholds are set at 0.55 for OpenNeRF and 0.5 for LeRF.
    \item \textbf{Loss Weights.} We balance the three proposed losses to optimize the trainings and obtain $\calL = w_{proj}\calL_{proj} + w_{irr}\calL_{irr} + w_{proto}\calL_{proto}$ with $w_{proj} = 20$, $w_{irr} = 1$ and $w_{proto} = 0.5$.
\end{itemize}

Finally, for all experiments, the model is trained for solely 200 epochs with an EMA decay factor $\alpha=0.998$ and an Adam optimizer of learning rate exponentially decreasing from $1e-2$ to $1e-4$ across the optimization.

\subsection{Ablative Experiments on USS (\autoref{sec:uss_exp})}
\label{sec:ablatives}
In the main paper, in order to propose a solution for the OV-SD problem adapted to Neural Fields, we began by proposing a novel USS NeRF-based method as, to the best of our knowledge, no existing USS method exist for the NeRF representation. In this section, we perform ablative experiments to evaluate the contributions of the different modules of our USS branch and show the results in \autoref{tab:uss_abl}. We evaluate the full method on USS (i.e. with no user query) and then either modify the projector (we test a simple linear MLP) or disable separately various components: the linear scheduling of $\beta$ and the two different ponderations of the prototypes update EMA process. \\
We note that the selected architecture used in DiSCO-3D indeed presents the best results amongst the different versions. Each of the other versions outputs diminished results, ranging from minimal loss of performances when changing the projector to maximal degradation when foregoing both ponderations in the EMA process. 
    

\begin{table*}[tb]
\centering
\begin{tabular}{cccc cc}
\hline
Decreasing $\beta$      &   Projector      &   Ponderation by $D_k$    &       Ponderation by $w_k$     & mIoU $\uparrow$  & mAcc $\uparrow$   \\ \hline

\cmark      &         Full        &       \cmark      &    \cmark  &  \textbf{27.47}    &  \textbf{ 51.99  }        \\
\cmark      &            2 Linear Layers       &       \cmark      &    \cmark  &   27.12   &    50.88      \\
\xmark      &           Full        &       \cmark      &    \cmark  & 26.52     &     50.41     \\


\cmark      &         Full        &       \xmark      &    \cmark  &   26.17  &   50.59     \\

\cmark      &             Full        &       \cmark      &    \xmark  &   26.02 &   50.09    \\

\cmark      &              Full        &       \xmark      &    \xmark  & 16.77     &    43.68     \\
\hline
\end{tabular}
\caption{\textbf{DiSCO-3D Ablative Experiments for USS.  }
}
\label{tab:uss_abl}
\end{table*}

\subsection{Open-Vocabulary Sub-concepts Discovery}

\textbf{Replica Sub-Concepts Dataset.} The complete list of groupings of our extended Replica dataset can be found in \autoref{tab:subconcept}. \\
\textbf{Naive Baselines Visualization.} In \autoref{sec:subconceptDiscovery}, we quantitatively compared DiSCO-3D with two naive baselines designed for the OV-SD problem. These baselines use the same architecture and configuration as DiSCO-3D but differ fundamentally in their segmentation process. Instead of jointly performing OV-Seg and USS as in our method, they execute the two tasks sequentially, each following a specific order.

\begin{figure}[t]
    \includegraphics[width=\columnwidth]{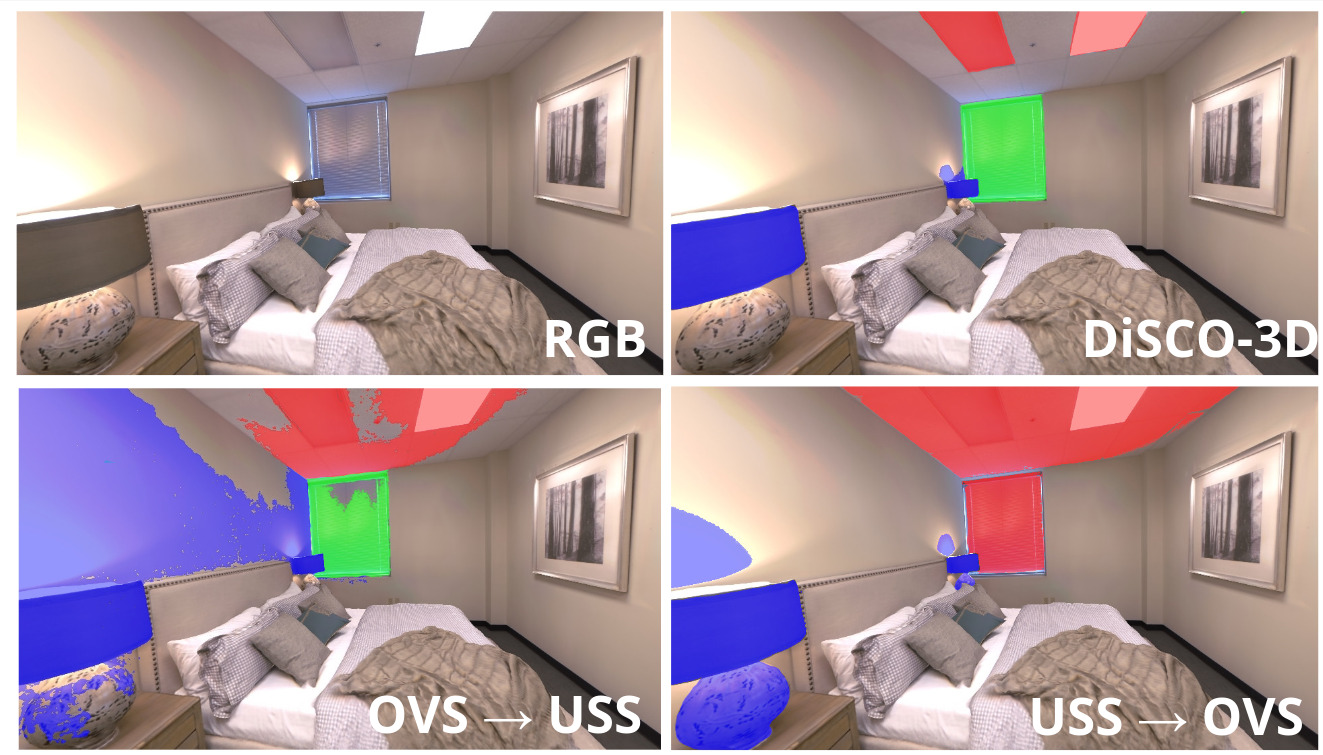}
    \caption{\textbf{OV-SD Example Naive Baselines vs DiSCO-3D.} The query is "light".}
    \label{fig:naive}
\end{figure}

We refer to these baselines as "naive" because a straightforward approach to solving OV-SD might be to apply OV-Seg and USS successively. However, as demonstrated in the quantitative evaluation presented in the main paper, this approach has notable shortcomings. To complement these results, Figure \autoref{fig:naive} provides a visual comparison using the query "light," which should correspond to the \textit{window}, the \textit{bed-side lamps} and the \textit{ceiling lamps}.

For the OVSeg-to-USS baseline, segmentation performance is significantly reduced due to the spatial imprecision of open-vocabulary relevancy. This leads to two key issues: (1) irrelevant objects may be partially segmented due to relevancy spilling (e.g., a large part of the wall above the bed), and (2) relevant objects, such as the window, may be incompletely segmented because the computed relevancy does not fully encompass the object. In contrast, DiSCO-3D mitigates these issues by leveraging DINO features as input to the projector, allowing it to refine spatial precision and avoid these relevancy errors.

For the USS-to-OVSeg baseline, while the segmentation aligns better with the scene’s geometry, the main issue lies in classification. Since USS is performed without query information, the resulting clusters are not structured according to the query. As a consequence, after OV-Seg filtering, objects that should be distinguished with respect to the query remain grouped together based on their overall similarity in the scene, leading to incorrect decomposition. Here, the ceiling with its lamps are segmented together with the window. Because the average CLIP embedding answers to the query, this grouping is considered a sub-concept in this naive baseline. \\
\textbf{Additional Results and Analysis.} We display in \autoref{tab:ovsd_abl1} and \autoref{tab:ovsd_abl2} additional metrics on the experiments of the main paper. We complete the given metrics by giving also the segmentation quality (SQ) and recognition quality (RQ) metrics, considered as sub-metrics of PQ such that :
\begin{equation}
    PQ = \underbrace{\frac{\sum_{(p, g)\in TP} \text{IoU}_{pg}}{|TP|}}_{SQ} \underbrace{\frac{|TP|}{|TP| + 0.5 |FP| + 0.5 |FN|}}_{RQ}
\end{equation}  

Finally, while we chose to display in the main paper the mIoU and mAcc metrics computed on the relevant classes, we complete the evaluation here by augmenting those metrics' computations with the irrelevant class. Note that because the background class presents better quantitative metrics in average due to the sheer size of the irrelevant class against the relevant sub-concepts, its metrics are much higher and thus might bias the global results towards the background class. 

\begin{figure}[tb]
    \includegraphics[width=\columnwidth]{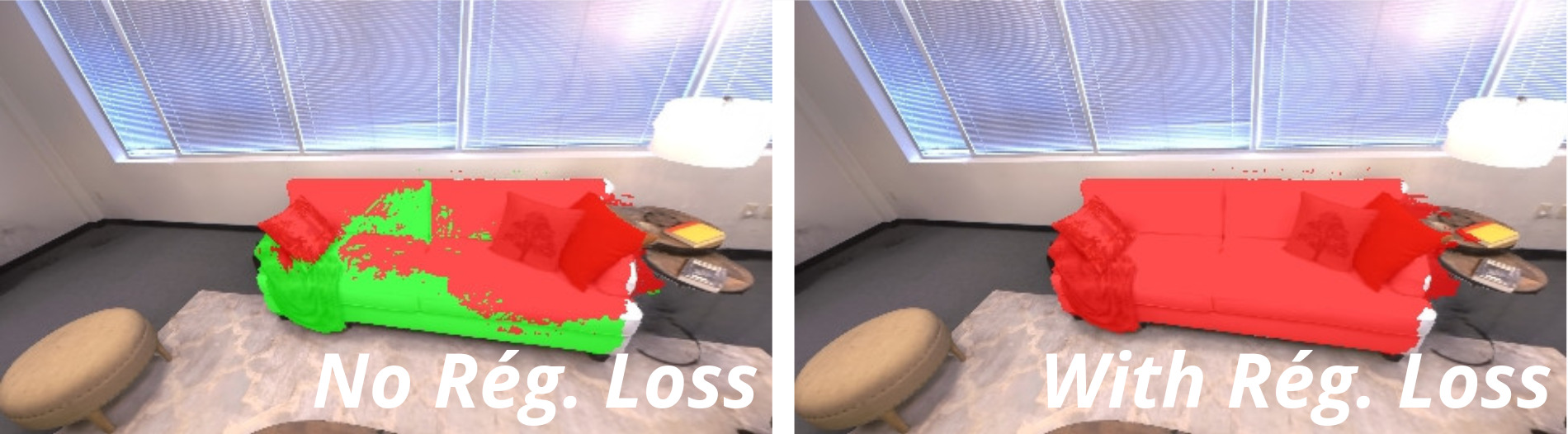}
    \caption{\textbf{Effect of the Regularization Loss.} Adding the regularization loss reduces over-segmenting (ie. describing single objects with more than one prototype). The query is "furniture".}
    \label{fig:proto_sup}
\end{figure}

\autoref{tab:proto_sup} gives additional metrics for the ablative experiment on the number of prototypes (\autoref{tab:proto} in the main paper) and \autoref{fig:proto_sup} shows a visual examples on how adding the regularization loss reduces over-segmenting of objects.

\begin{figure*}[t!]
    \includegraphics[width=\textwidth]{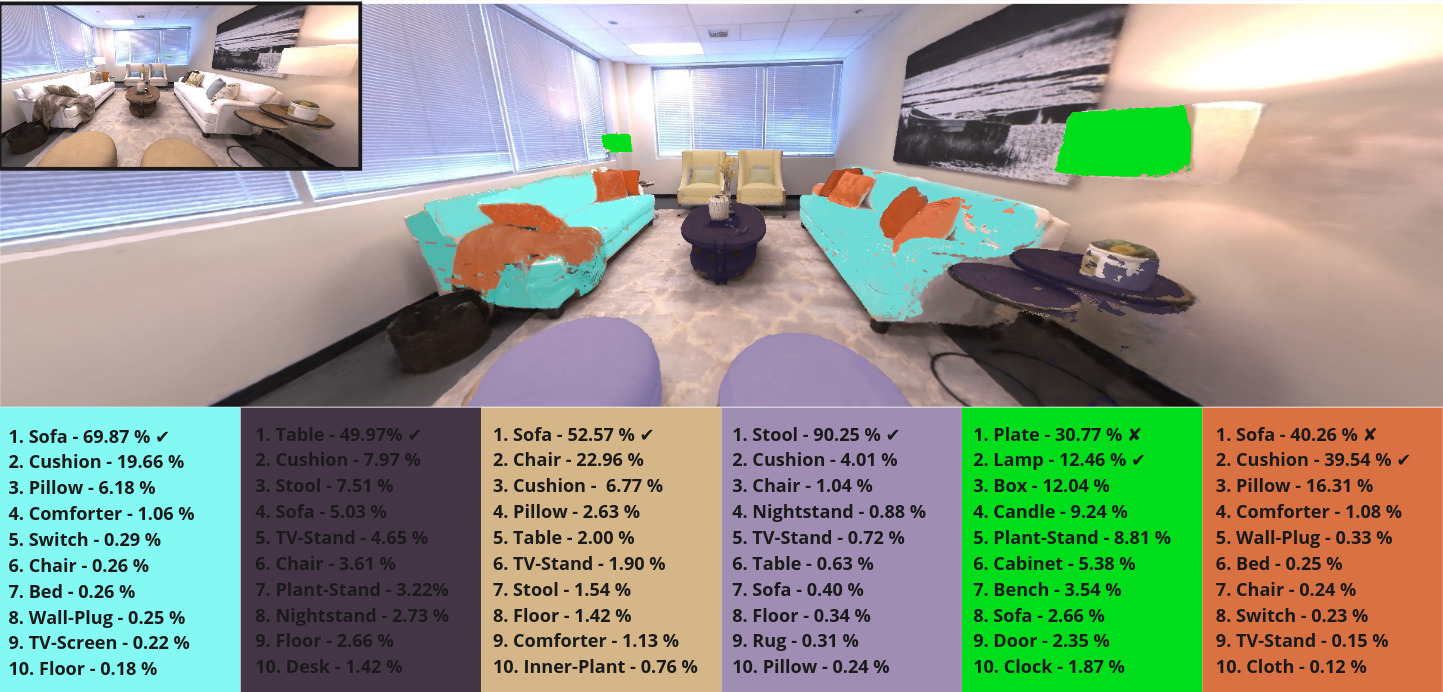}
    \caption{\textbf{Top-10 Class Labels Linking for every Sub-Concepts.} The query is "furniture".}
    \label{fig:proto_abl}
\end{figure*}



\begin{table}[tb]
\centering
\scalebox{0.73}{
    \begin{tabular}{c | c | c | c | c | c | c || c}
$\calL_{proto}$ & $N_{add}$  & 0 & 2 & 5 & 10 & 20 & $N=10$\\\hline
 \multirow{4}{*}{\ding{51}} & Used $N_{add}$ & -0.12 & 1.08 & 1.52 & 1.91 & 1.96 & 1.80 \\ 
& PQ $\uparrow$  &  8.53 & 9.52  & 10.06  & 10.15  & 10.12  &  \textbf{10.19} \\
&  mIoU $\uparrow$  & 8.81 & 10.45  & 12.38  & 12.70  & 12.59  &  \textbf{12.77} \\
&  mAcc $\uparrow$  &  36.72 & 39.60  & 42.81  & 43.63  & 43.47  &  \textbf{44.29} \\ \hline
 
 \multirow{4}{*}{\ding{55}} & Used $N_{add}$ & -0.07 & 1.33 & 1.98 & 2.62 &3.02 & 2.60 \\ 
  & PQ $\uparrow$  &  8.56 & 9.49  & 9.72  & 9.71  & 9.55  &  \textbf{9.77} \\
  &  mIoU $\uparrow$  & 8.77 & 10.27  & 12.13  & \textbf{12.42}  & 12.30  &  12.35 \\
  &  mAcc $\uparrow$  &  35.82 & 39.06  & 42.52  & 43.14  & 42.64  &  \textbf{43.36} \\
 
    \end{tabular}
}
\caption{\textbf{Additional metrics on the ablative on $\#$ of Prototypes} ($N = N_{GT} + N_{add}$).}
\label{tab:proto_sup}
\end{table}

\textbf{Another example of using the CLIP Prototypes.}
In figure \autoref{fig:resuLabel} of the main article, we show results of \textit{a posteriori} linking of the automatic sub-concepts with class names using the corresponding CLIP prototypes. Here, we dive deeper and provide another example in \autoref{fig:proto_abl} where we give the corresponding probability attributions of the top-10 semantic classes (amongst the 51) of each sub-concept. We query the scene for "furniture" and compute for each CLIP prototype the distances to each CLIP embedding of the 51 semantic classes. The probability distribution is then obtained by performing a softmax operation on the inverse of the distances (multiplied by a factor 100 to accentuate the sharpness of the distribution, as the distances between an image CLIP embedding and a text CLIP embedding are all rather close). Although 2 out of the 6 sub-concepts are not linked to the correct semantic classes, the 4 other correct classes are predicted with high confidence (up to $90.25\%$ for the "stool" sub-concept class), showing the confidence of our model with unambiguous concepts. Regarding the incorrect predictions, we can first notice that the cushion class is the second most probable prediction with only $0.72\%$ of differences in confidence. This result reflects that the associated CLIP prototype refers to an intermediate concept corresponding to a sofa-cushion, a cushion in itself not being a furniture while a cushion as part of a sofa can be considered as one. Similarly, the CLIP prototype corresponding to the armchair matches with the sofa at 52.57\% and with a chair at 22.96\%. This is consistent with the definition of an armchair: an intermediate concept between a sofa and a chair.
The other incorrect sub-concept corresponding to the lamp is the less correct prediction, as once again the second probable prediction but with more differences in confidence. However, this error can be partly explained by the difficulty of the prediction as the lamp object in itself has a peculiar form less discriminative than the form of a sofa. 





\begin{figure}[tb]
    \includegraphics[width=\columnwidth]{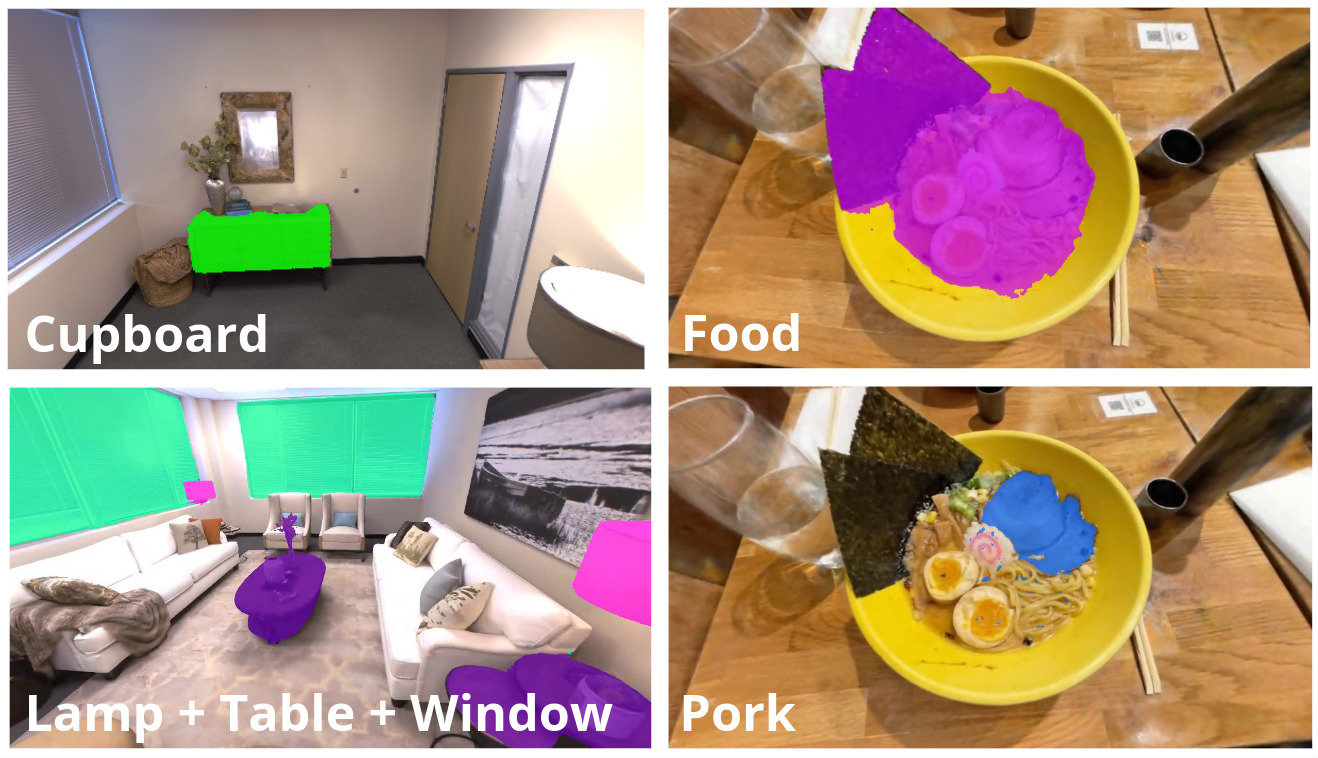}
    \caption{\textbf{Additional OVSeg Results.}}
    \label{fig:ovs}
\end{figure}

\begin{figure}[tb]
    \includegraphics[width=\columnwidth]{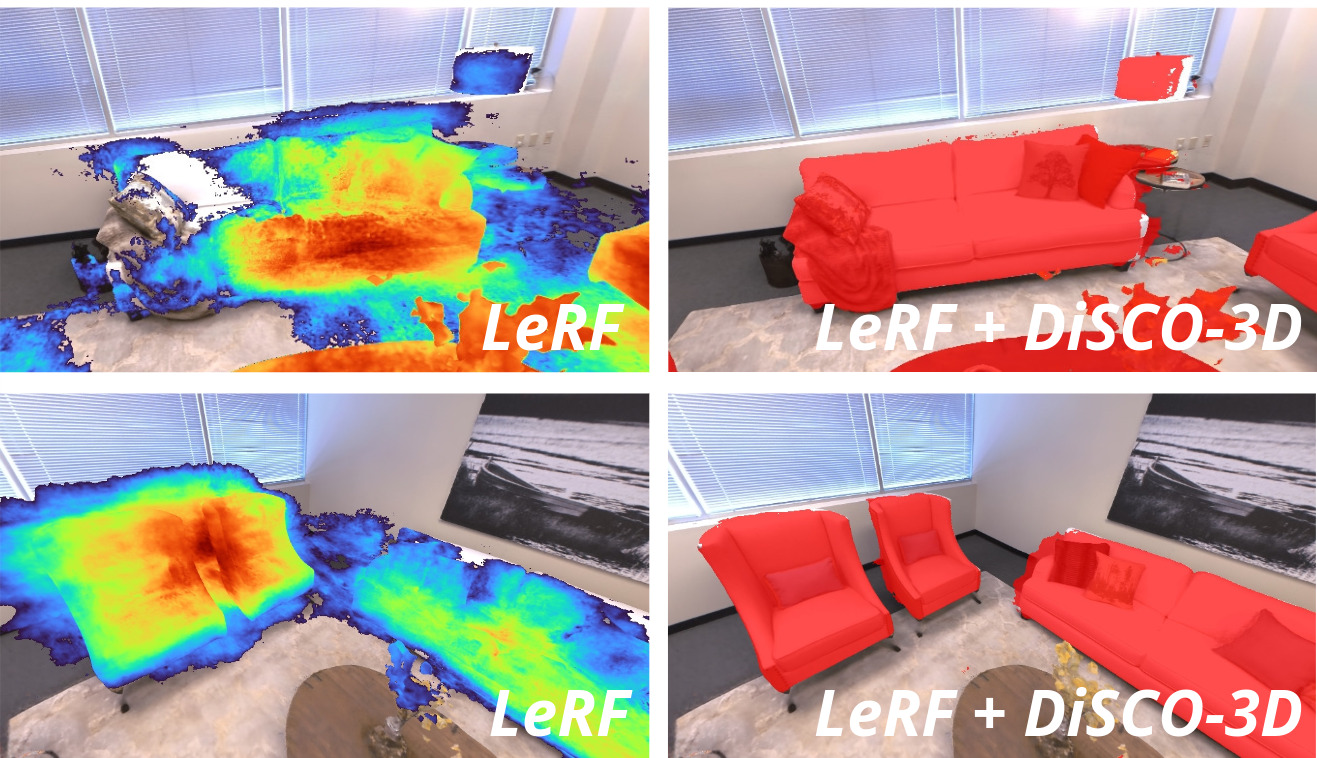}
    \caption{\textbf{Display of relevancy holes and spillings.} The queries of the fist and second lines are respectively "Furniture" and "Seatings" in the OV-Seg setting.}
    \label{fig:spill}
\end{figure}

\subsection{Open-Vocabulary Segmentation}
\label{sec:ovseg_abl}

We display some additional qualitative results in \autoref{fig:ovs}, both with singular queries (composed of precise classes and concepts) and multiple queries at once. 

\textbf{Relevancy Holes and Spilling.} We stated in \autoref{sec:ovseg} that DiSCO-3D is able to mitigate common open-vocabulary segmentation issues, namely relevancy spillings and relevancy holes. We illustrate this affirmation in \autoref{fig:spill}. Relevancy holes, displayed on the first lines, define areas where only parts of an object relevant to the query in theory responds well in practise. DiSCO-3D succeeds in completing the segmentation to encompass the whole object in the segmentation. Relevancy spilling rather relates to the opposite phenomenom, where irrelevant areas around a relevant object can be detected by the open-vocabulary models due to spatial imprecision. This is illustrated in the second line of the figure. DiSCO-3D also reduces this issue by focusing only on highly relevant areas and completing them.

\textbf{Mono-Label Paradigm.} Additionally to what we call the \textit{multi-label} paradigm (ie. each 3D point can be assigned zero or multiple labels based on independent query predictions), some methods such as \cite{engelmann2024opennerf} evaluate themselves on the \textit{mono-label} paradigm where each point receives a single label corresponding to the most probable class amongst all queries. Regarding DiSCO, this translates into training $1$ models with $N_q$ simultaneous queries, meaning that this paradigm is a way to evaluate DiSCO's ability to handle multiple queries at once. Note that because this paradigm needs the labels to be non-overlapping and needs to cover the whole scene, it cannot be evaluated on the grouping dataset which has overlapping queries (eg. "furnitures" and "seating" have common sub-concepts). The \textit{multi-label} setup is considered more challenging as it requires segmenting each class independently without relying on other class names as priors. \\
We report quantitative results of this paradigm in \autoref{tab:open-seg-abl}. Regarding LeRF, we notice improvements for every metrics and paradigms when adding DiSCO ($+2.94$ mIoUs and $+6.35$ mAccs respectively). This is because applying DiSCO to LeRF greatly improves the segmentation by reducing the relevancy spilling (as illustrated in \autoref{fig:limit1}): it directly improves mIoU and also increases mAcc because reducing spilling in a paradigm where every point is labeled increases correct classification. 
Regarding OpenNeRF, whose segmentation performances are already much better than LeRF, integrating DiSCO slightly improves the mIoUs (resp. $+1.68$) at the expense of mAcc. This trade-off arises because DiSCO segments directly from features rather than relying on similarity maps like OpenNeRF. As a result, DiSCO provides better boundary refinement by leveraging additional information but introduces minor misclassification, particularly for small less frequent classes.

\begin{table}[tb]
\centering
\scalebox{0.81}{
    \begin{tabular}{l ll }
    \hline
        \multirow{2}{*}{Method} &
          \multicolumn{2}{c}{\textit{Mono-Label}} 
          \\ \cline{2-3}
        & mIoU $\uparrow$  & mAcc $\uparrow$  \\ 
        \hline
    
    LeRF~\cite{kerr2023lerf}       & \doriand{ 10.49}    &  \doriand{22.02}    \\ 
    LeRF + DiSCO-3D  &   \doriand{\textbf{13.43}}   &  \doriand{\textbf{28.37}}    \\ \hline 
    OpenNeRF~\cite{engelmann2024opennerf} &  \doriand{19.08}    &  \doriand{\textbf{31.96}} \\ 
    OpenNeRF + DiSCO-3D  &  \doriand{\textbf{20.76}}    &  \doriand{30.19}     \\ \hline
    \end{tabular}
}
\caption{\textbf{DiSCO-3D Quantitative Evaluation for OV-Seg in the \textit{mono-label} paradigm.}}
\label{tab:open-seg-abl}
\end{table}

\begin{figure}[b]
    \includegraphics[width=\columnwidth]{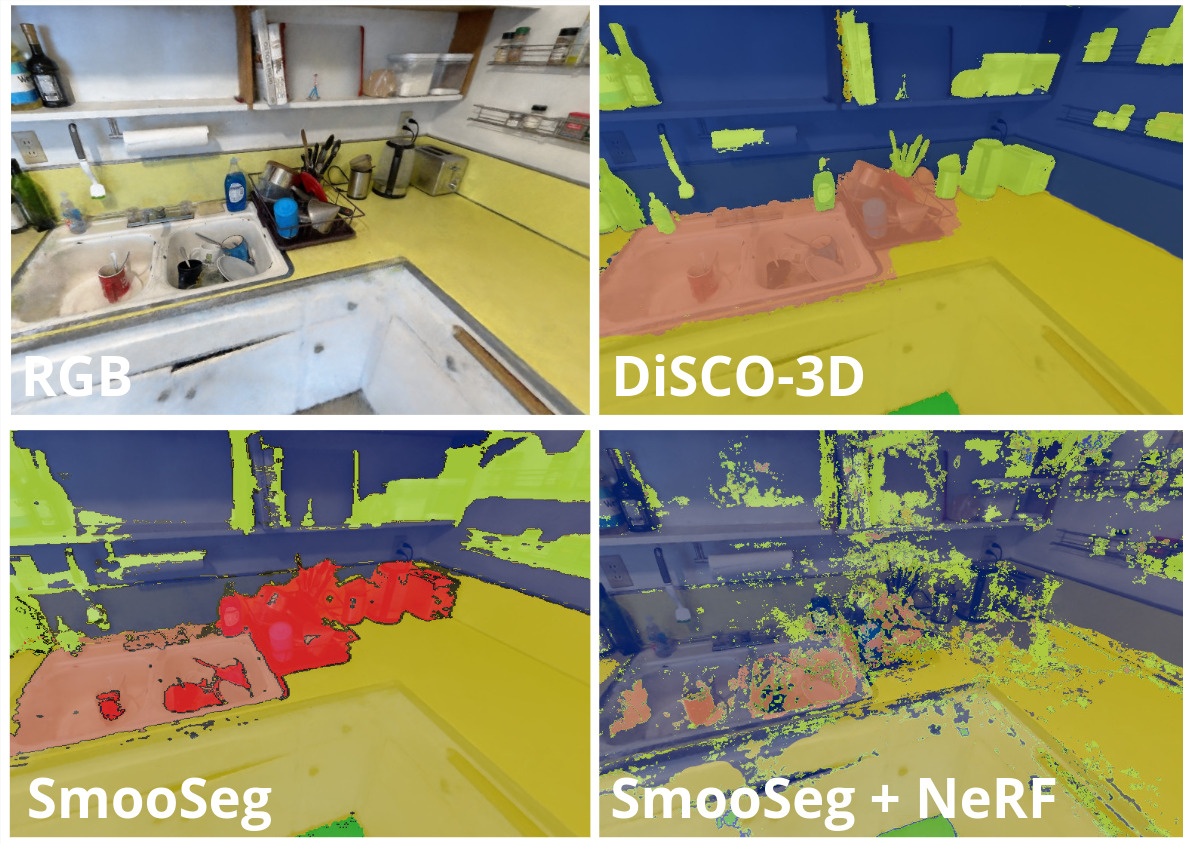}
    \caption{\textbf{Example of USS on real data. \stevermk{Plus gros, avec imagette sans segmentation pour qu'on voit à quoi ressemble la scène}}}
    \label{fig:kitchen}
\end{figure}

\begin{figure}[b]
    \includegraphics[width=\columnwidth]{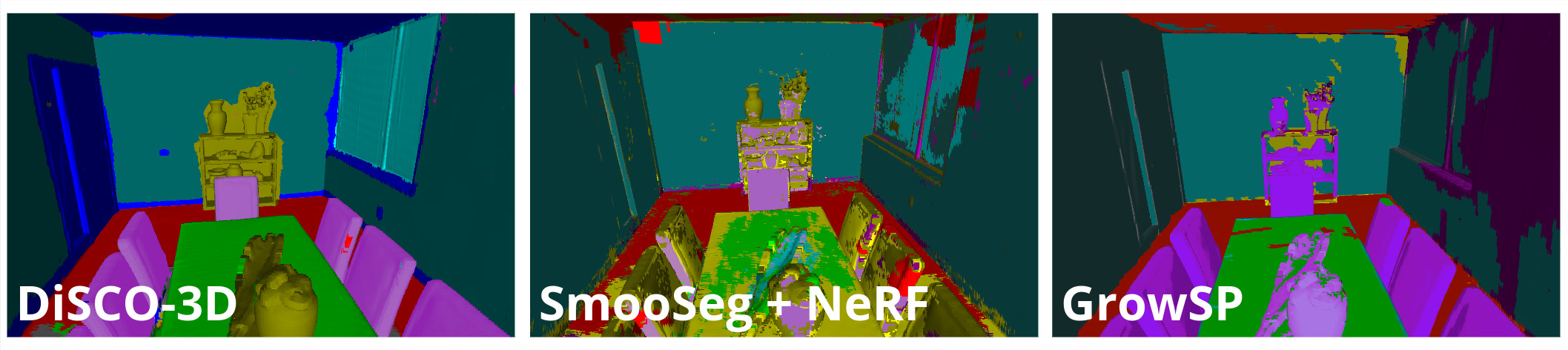}
    \caption{\textbf{USS on the 3D Point Cloud of Replica.}}
    \label{fig:uss_abl}
\end{figure}

\subsection{Unsupervised Semantic Segmentation}
\label{sec:uss_abl}
We display here two figures of 3D USS. \autoref{fig:kitchen} shows an example on real 2D data (in particular the "Waldo Kitchen" scene from LeRF). On this latter figure, the "SmooSeg" baseline refers to the 2D method being trained on the multi-view images without injecting 3D inside the segmentation, while the "SmooSeg + NeRF" image is a semantic render from a NeRF model trained using the segmentation maps of the SmooSeg. As explained in \autoref{sec:uss_exp}, 2D USS methods such as SmooSeg do not have multi-view consistency. This makes the training of a Semantic-NeRF hardly consistent, resulting in very noisy segmentations. Although multi-view inconsistent, SmooSeg actually performs well when doing per-image segmentation as illustrated in the figure. While some noise subsist, the results are semantically and spatially coherent. However, DiSCO-3D still produces better segmentation as it profits from multi-view information for more precise DINO features, thus better spatial precision of the segmentation (e.g. the bottles on the top left of the image). Note that no GrowSP results can be obtained as there is no available point cloud for these hand-captured images of real data. Similarly, figure \autoref{fig:uss_abl} displays 3D point cloud segmentation results (used for quantitative evaluation) on Replica. The obtained renders are consistent with previous observations, as SmooSeg lacks multi-view consistency once again. Although GrowSP gives better segmentation with actually more precise details (e.g. the background shelves) but there are several areas with unexpected spillings which degrades the segmentation.













\begin{table*}[t]
\centering
\begin{tabular}{l l l}

    \hline
    \textbf{ID} & \textbf{Concept} & \textbf{Associated Sub-Concepts} \\
    \hline
    1 & Furniture & chair, sofa, bench, stool, table, desk, cabinet, nightstand, shelf \\
    2 & Seating & chair, sofa, bench, stool, cushion, pillow \\
    3 & Sleeping & bed, comforter, blanket, pillow \\
    4 & Storage & cabinet, shelf, basket, box, desk-organizer \\
    5 & Walls & wall, panel \\
    6 & Floors & floor, rug \\
    7 & Ceilings & ceiling, vent \\
    8 & Entrances & door, window, blinds \\
    9 & Screens & tv-screen, monitor, tablet \\
    10 & Light & lamp, candle \\
    11 & Plants & indoor-plant, plant-stand \\
    12 & Art & picture, sculpture \\
    13 & Time & clock \\
    14 & Trash & bin \\
    15 & Soft & pillow, cushion, comforter, blanket, bed, cloth \\
    16 & Decor & sculpture, vase, candle \\
    17 & Organize & desk-organizer, box, basket \\
    18 & Airflow & vent \\
    19 & Work & desk, monitor, lamp \\
    20 & Eat & table, plate, bowl \\
    21 & Reflect & monitor, tv-screen \\
    22 & Warm & blanket, cloth \\
    23 & Watch & tv-screen, monitor, tablet \\
    24 & Tidy & desk-organizer, basket \\
    25 & Walk & floor, rug \\
    26 & Container & pot, bottle \\
    27 & Press & switch \\
    28 & Cushion & cushion, pillow \\
    29 & Displays & tv-screen, monitor, tablet \\
    30 & Rest & sofa, bed, pillow \\
    31 & Relax & sofa, chair, bed, cushion, pillow, blanket \\
    32 & Electronics & monitor, tablet, tv-screen, clock, camera \\
    33 & Lounge & sofa, bench, pillow, cushion \\
    34 & Dining & table, plate, bowl, bottle \\
    35 & Ventilation & vent, window \\
    36 & Opening & door, window, blinds \\
    37 & Comfort & pillow, cushion, blanket, bed, sofa \\
    38 & Portable & basket, box, tablet \\
    39 & Fragile & vase, sculpture, monitor, tv-screen \\
    40 & Heavy & table, cabinet, sofa, bed, sculpture \\
\hline
\end{tabular}
\caption{\textbf{Replica Sub-Concepts Dataset.}
}
\label{tab:subconcept}
\end{table*}


\clearpage

\begin{table*}[tb]
\centering
\scalebox{0.8}{
    \begin{tabular}{@{\extracolsep{8pt}}ll lll llll@{}}
    \hline
        
       \multirow{2}{*}{FF} & \multirow{2}{*}{Method}  &  
          \multicolumn{7}{c}{$\calP_{CLIP}$} 
           \\ \cline{3-9} 
      &  & PQ $\uparrow$  & RQ $\uparrow$ & SQ $\uparrow$ & $\textrm{mIoU}_{rel} \uparrow$ & $\textrm{mAcc}_{rel} \uparrow$ & $\textrm{mIoU}_{all} \uparrow$ & $\textrm{mAcc}_{all}$ 
        \\ 
        \hline
    
    \multirow{3}{*}{\rotatebox[origin=c]{90}{LeRF}} &   USS → OVS      &  4.76   &  32.48  & 11.62  &  6.52 & 22.54 & 29.89 & 44.12     \\
    &   OVS → USS      &   5.99  &  30.47  & 13.41  & 8.71 & 21.44 & 39.82 & 49.38    \\
    &    DiSCO-3D  &  \textbf{8.13}  & \textbf{45.45}   & \textbf{15.39} & \textbf{10.79} & \textbf{33.39} & \textbf{40.64} & \textbf{58.58}    \\ \hline 
     \multirow{3}{*}{\rotatebox[origin=c]{90}{\footnotesize OpenNeRF}} & USS → OVS &  4.97   & 25.01  & 13.02 & 6.08    & 13.98 & 30.44 & 39.71   \\ 
     & OVS → USS &  5.47   & 24.11  & 13.40   & 8.94 & 13.56 & 38.66 & 41.99    \\ 
    &  DiSCO-3D  &  \textbf{8.65}  & \textbf{39.36} & \textbf{17.84} & \textbf{10.82} & \textbf{19.24} & \textbf{40.57} & \textbf{49.88}    \\ \hline
    \end{tabular}
}
\caption{\textbf{DiSCO-3D Quantitative Evaluation for OV-SD using $\calP_{CLIP}$ matching.} }
\label{tab:ovsd_abl2}
\end{table*}

\begin{table*}[tb]
\centering
\scalebox{0.8}{
    \begin{tabular}{@{\extracolsep{8pt}}ll lll llll@{}}
    \hline
        
       \multirow{2}{*}{FF} & \multirow{2}{*}{Method}  &  \multicolumn{7}{c}{\textit{Hungarian}} 
           \\ \cline{3-9} 
      &  & PQ $\uparrow$  & RQ $\uparrow$ & SQ $\uparrow$ & $\textrm{mIoU}_{rel} \uparrow$ & $\textrm{mAcc}_{rel} \uparrow$ & $\textrm{mIoU}_{all} \uparrow$ & $\textrm{mAcc}_{all} \uparrow$ 
        \\ 
        \hline
    
    \multirow{3}{*}{\rotatebox[origin=c]{90}{LeRF}} &   USS → OVS      &  6.94   & 53.96   & 11.72  &  10.92 & 35.57 & 34.70 & 55.60      \\
    &   OVS → USS      &  7.48 & 44.09 & 13.24  &  10.90 & 27.11 & 41.50  & 54.74    \\
    &    DiSCO-3D  & \textbf{10.19}   & \textbf{57.54}   & \textbf{14.64} &  \textbf{12.77} & \textbf{44.29}  & \textbf{42.61} & \textbf{63.49} \\ \hline 
     \multirow{3}{*}{\rotatebox[origin=c]{90}{\footnotesize OpenNeRF}} & USS → OVS &  6.53 & 38.29 & 12.77 & 8.67 & 23.85 & 38.52 & 52.54   \\ 
     & OVS → USS &  6.73   & 34.84  & 13.31   & 10.58 & 22.00 & 41.72 & 51.86      \\ 
    &  DiSCO-3D  & \textbf{10.49}   & \textbf{52.42} & \textbf{16.65} & \textbf{12.69} & \textbf{29.06} & \textbf{42.23} & \textbf{55.82}      \\ \hline
    \end{tabular}
}
\caption{\textbf{DiSCO-3D Quantitative Evaluation for OV-SD using Hungarian Matching.} }
\label{tab:ovsd_abl1}
\end{table*}

\end{document}